\documentclass{article}

\usepackage{PRIMEarxiv}

\usepackage[utf8]{inputenc} 
\usepackage[T1]{fontenc}    
\usepackage{hyperref}       
\usepackage{url}            
\usepackage{booktabs}       
\usepackage{amsfonts}       
\usepackage{nicefrac}       
\usepackage{microtype}      
\usepackage{lipsum}
\usepackage[pdftex]{graphicx}
\graphicspath{{media/}}     
\usepackage{multirow}

\title{Region-CAM: Towards Accurate Object Regions in Class Activation Maps for Weakly Supervised Learning Tasks
}

\author{
  Qingdong Cai \\
  School of Electrical and Electronic Engineering\\
  The University of Sheffield\\
  Sheffield\\
  \texttt{qcai7@sheffield.ac.uk} \\
   \And
  Charith Abhayaratne \\
  School of Electrical and Electronic Engineering \\
  The University of Sheffield \\
  Sheffield\\
  \texttt{c.abhayaratne@sheffield.ac.uk} \\
}

\begin{document}
\maketitle

\begin{abstract}
Class Activation Mapping (CAM) methods are widely applied in weakly supervised learning tasks due to their ability to highlight object regions. However, conventional CAM methods highlight only the most discriminative regions of the target. These highlighted regions often fail to cover the entire object and are frequently misaligned with object boundaries, thereby limiting the performance of downstream weakly supervised learning tasks, particularly Weakly Supervised Semantic Segmentation (WSSS), which demands pixel-wise accurate activation maps to get the best results. To alleviate the above problems, we propose a novel activation method, Region-CAM. Distinct from network feature weighting approaches, Region-CAM generates activation maps by extracting semantic information maps (SIMs) and performing semantic information propagation (SIP) by considering both gradients and features in each of the stages of the baseline classification model. Our approach highlights a greater proportion of object regions while ensuring activation maps to have precise boundaries that align closely with object edges. Region-CAM achieves 60.12\% and 58.43\% mean intersection over union (mIoU) using the baseline model on the PASCAL VOC training and validation datasets, respectively, which are improvements of 13.61\% and 13.13\% over the original CAM (46.51\% and 45.30\%). On the MS COCO validation set, Region-CAM achieves 36.38\%, a 16.23\% improvement over the original CAM (20.15\%). Moreover, when the class activation methods of other WSSS algorithms are replaced with Region-CAM, the accuracy of the segmentation seed generated by these algorithms is further improved. We also demonstrate the superiority of Region-CAM in object localization tasks, using the ILSVRC2012 validation set. Region-CAM achieves 51.7\% in Top-1 Localization accuracy (\textit{Loc1}). Compared with LayerCAM, an activation method designed for weakly supervised object localization, Region-CAM achieves 4.5\% better performance in \textit{Loc1}.
\end{abstract}


\section{Introduction}
\label{sec:introduction}
The Class Activation Mapping (CAM) method was originally developed to provide a visual explanation of Convolutional Neural Network (CNN) predictions by highlighting image regions that significantly contribute to the decisions of model~\cite{zhou2016learning}. Since CAM methods can highlight target object regions by generating activation maps from classification networks \cite{selvaraju2017grad,wang2020score}, in which pixels with higher activation values indicate a greater likelihood of belonging to the target object, they are widely employed in tasks like Weakly Supervised Semantic Segmentation (WSSS) \cite{zhang2020survey,li2018weakly,li2023weakly,ru2022weakly,alom2018recurrent,chen2017deeplab,chen2018encoder,cheng2022masked} and Weakly Supervised Object Localization (WSOL) \cite{xu2022cream,choe2020evaluating,zhang2021weakly,kim2021normalization}, where pixel-wise labels are not available for training.

\begin{figure*}[t]
    \centering
    \begin{minipage}{0.1\textwidth} 
            \centering
            \includegraphics[width=\linewidth,height=7cm]{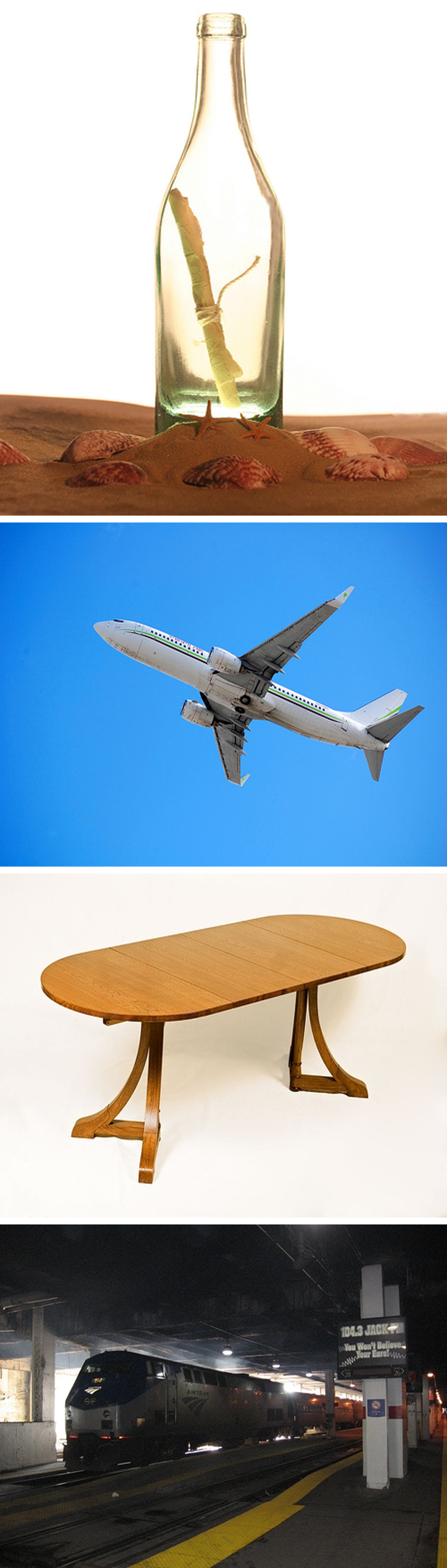}\\ 
            {(a)}\\
            {Input}
    \end{minipage}
    \begin{minipage}{0.39\textwidth}
        \centering
        \includegraphics[width=\linewidth,height=7cm]{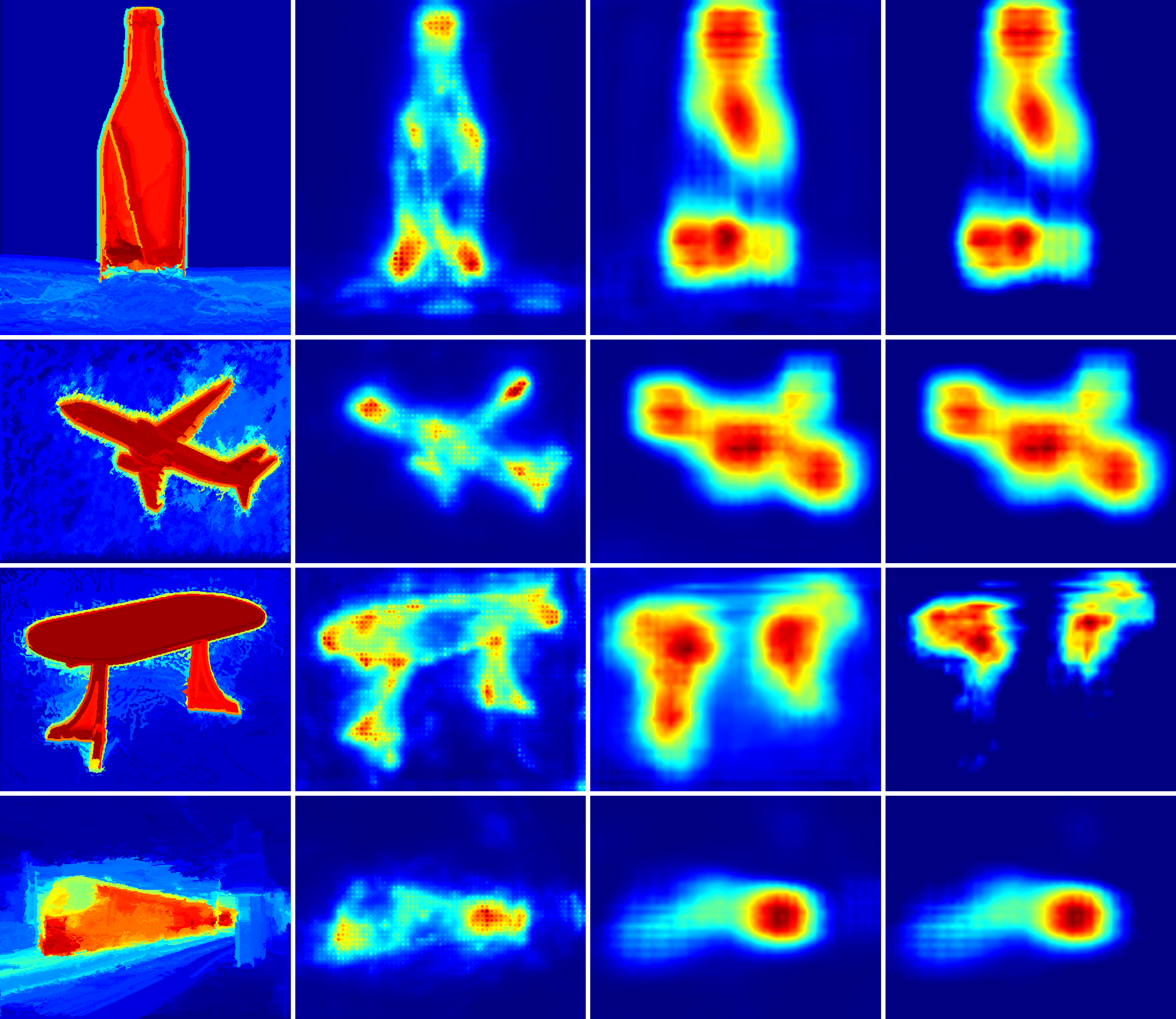}\\ 
        {(b)\hspace{0.18\textwidth} (c)\hspace{0.18\textwidth} (d)\hspace{0.18\textwidth} (e)}\\
        {Class activation maps}
    \end{minipage}
    \begin{minipage}{0.39\textwidth}
        \centering
        \includegraphics[width=\linewidth,height=7cm]{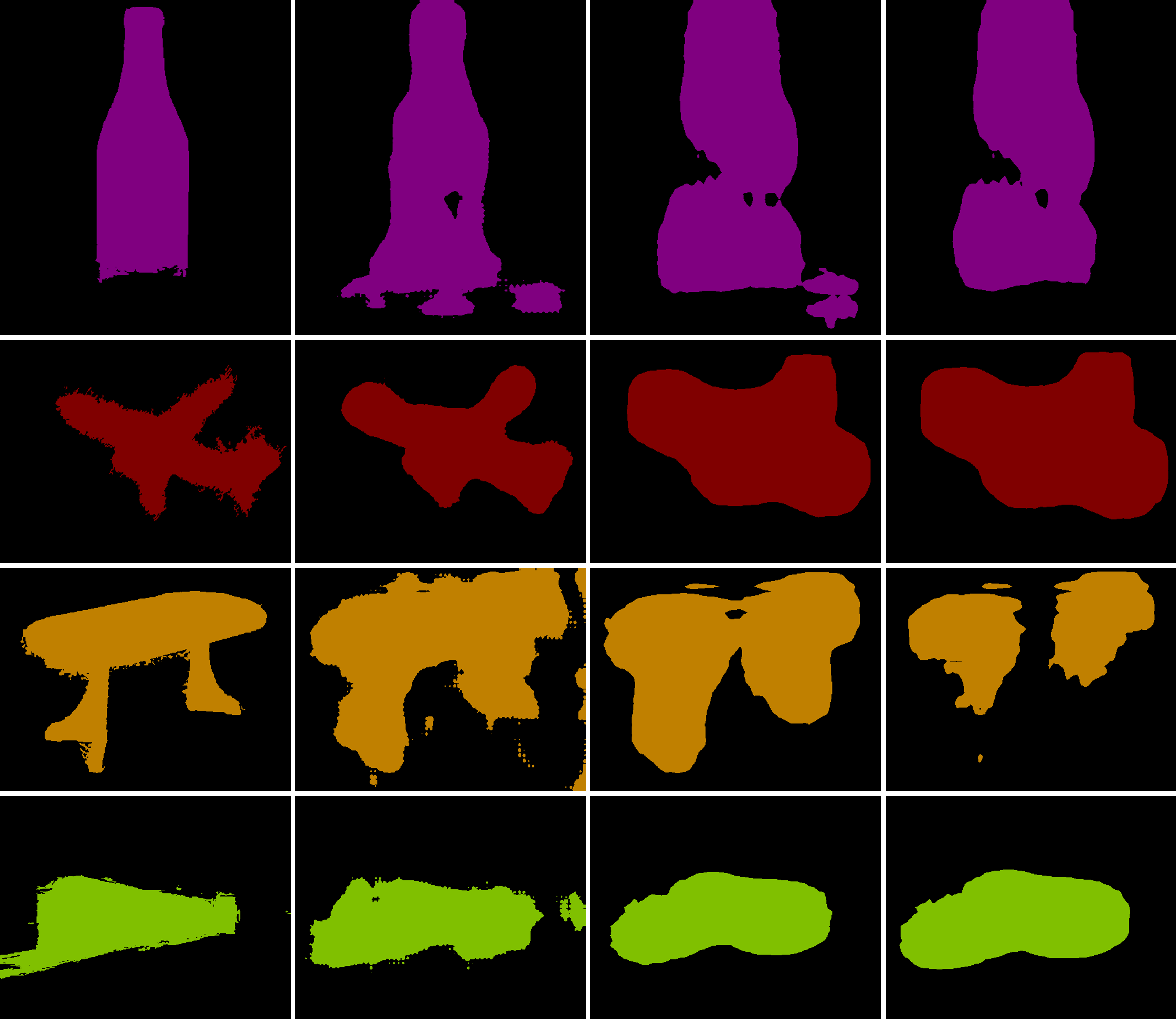}\\ 
        {(f)\hspace{0.18\textwidth} (g)\hspace{0.18\textwidth} (h)\hspace{0.18\textwidth} (i)}\\ {Segmentation seeds}
    \end{minipage}
    \begin{minipage}{0.1\textwidth}
        \centering
        \includegraphics[width=\linewidth,height=7cm]{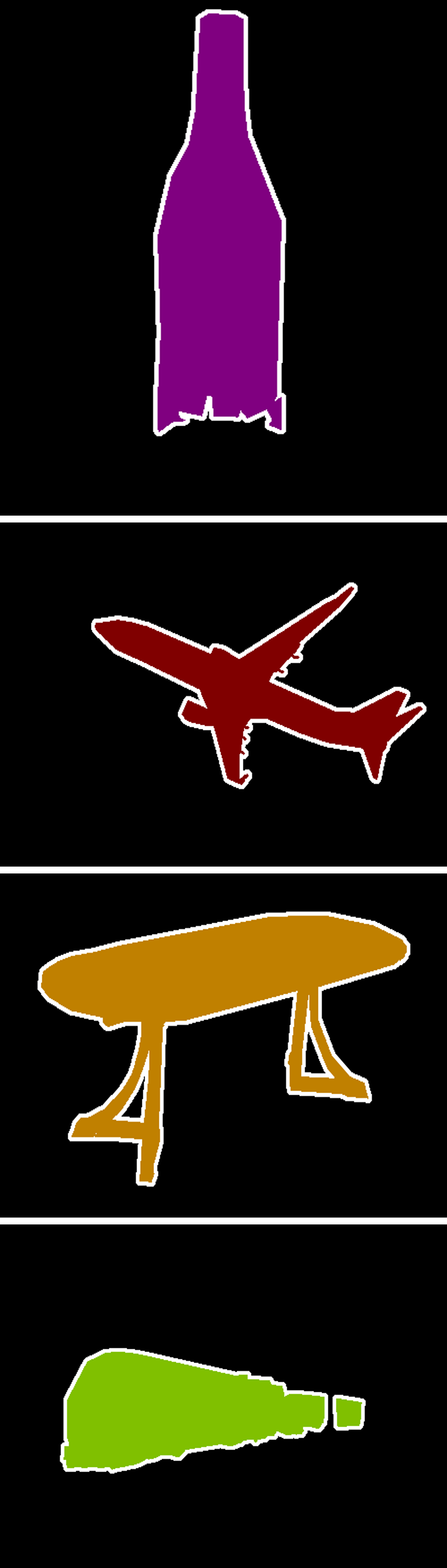}\\ 
        (j)\\{GT}
    \end{minipage}
    \caption{\textbf{Comparison of class activation maps and their derived semantic segmentation seeds for the current CAM methods and the proposed Region-CAM. Column (a): The input images; Columns (b)-(e): The activation maps are generated using different activation methods - Region-CAM (our proposed), LayerCAM, Grad-CAM, and CAM, respectively; Columns (f)-(i): The derived semantic segmentation seeds based on the activation maps in columns (b)-(e), respectively: Column (j): The Ground Truth (GT) masks.}}
    \label{Mycamresults}
\end{figure*}

Specifically, in image-level WSSS \cite{wang2020weakly,cai2023ssdb,wang2017weakly}, the primary research focus lies in achieving semantic segmentation using only classification labels as supervision signals \cite{shimoda2020weakly,yang2025exploring,chen2022class}. However, image-level supervision provides only coarse information about the categories present in an image, without specifying critical details such as the precise location, boundaries, shape, or size of the objects \cite{chou2017framecnn,wu2025prompt}. This lack of spatial annotation poses a significant challenge for achieving semantic segmentation \cite{wang2017weakly}. CAM methods address this gap by generating activation maps that highlight class-related regions and provide approximate localization cues \cite{ru2022weakly}. These activation maps are further transformed into segmentation seeds, which serve as supervisory labels for subsequent network training \cite{xu2019improved,zhang2020self}. However, widely applied activation methods \cite{zhou2016learning,selvaraju2017grad,chattopadhay2018grad}, often emphasize the most discriminative regions of the object class, since the activation maps are generated from the final convolutional layer of a CNN. These activated regions often fail to cover the entire object and are misaligned with the object boundary. As a result, the segmentation seeds derived from these activation maps are limited in accuracy, which affects the performance of WSSS methods \cite{ahn2019weakly,ahn2018learning,chen2020weakly}. Figure~\ref{Mycamresults} shows the class activation maps produced by CAM methods commonly used in the state-of-the-art WSSS segmentation methods (LayerCAM~\cite{jiang2021layercam}, Grad-CAM~\cite{selvaraju2017grad} and CAM~\cite{zhou2016learning}) and our proposed Region-CAM followed by the corresponding segmentation seeds derived from them, respectively for four sample images. As can be seen from Figure~\ref{Mycamresults}, the existing CAM methods lead to in inaccurate segmentation leads compared to the Ground Truth (GT) masks.

Previous studies \cite{jiang2021layercam,shi2021zoom,rebuffi2020there,yasuki2024cam,omeiza2019smooth} have realized the limitation of generating activation maps solely from the last convolutional layer of the network. To address this issue, they combined intermediate layer features to capture finer-grained object information~\cite{jiang2021layercam,shi2021zoom,rebuffi2020there,omeiza2019smooth}
or adopting a network with a large effective receptive field~\cite{yasuki2024cam}. However, these methods still lack the ability to generate accurate pixel-level activation maps. For instance, Zoom-CAM~\cite{shi2021zoom} refines activation maps by applying scalar weighting to intermediate-layer feature maps, which enhances fine-grained detail extraction, but it yields only marginal performance improvements. LayerCAM~\cite{jiang2021layercam} is able to activate a wider object region and capture more precise details compared to these methods, but the activation region is discontinuous, and multiple high-value points appear in the same object area. While such activation results can improve more accurate object localization results, their contribution to the WSSS field remains limited, as the segmentation seeds generated from this activation map still lacks sufficient target information. In addition, Yasuki and Taki~\cite{yasuki2024cam} combined large-core CNN, CAM, and simple data augmentation methods to obtain more comprehensive activation maps. However, their method requires data augmentation to be applied during network training, which may affect algorithm performance when their method is combined with other existing weakly supervised methods. Moreover, these methods~\cite{jiang2021layercam,yasuki2024cam} are designed for WSOL rather than for WSSS which emphasizes pixel-level accuracy. 

In order to generate fine-grained activation maps to benefit weakly supervised domain, we propose a novel activation method, Region-CAM in this paper. We develop our approach based on one hypothesis: we assume that points within the same superpixel region should contain the same semantic information, therefore the activation values of points within the same superpixel region should be the same. Moreover, we found that by performing a simple K-mean on the shallow features, object boundary and shape prior information can be captured. Based on this assumption and finding, our activation method considers gradients and features separately instead of using gradients to weight features like previous methods \cite{zhou2016learning,chattopadhay2018grad,selvaraju2017grad,jiang2021layercam,wang2020score}. We regard the non-negative gradients from different layers as Semantic Information Maps (SIMs), where the value at each point represents its importance to the target category \cite{chattopadhay2018grad}. These information maps are fused and further averaged over superpixel regions based on our designed Semantic Information Propagation (SIP) mechanism to obtain activation maps. Our activation method can be flexibly applied to different network structures and the activated target regions are more complete and accurate. We validate Region-CAM on various datasets, and the results show that our method activates more complete and accurate target regions than previous class activation methods. Moreover, Region-CAM can be flexibly integrated with WSSS methods to achieve better performance. The main contributions of our work are summarized as follows:
\begin{itemize}
    \item A novel activation method is presented that, in contrast to previous feature weighting approaches, considers gradients and feature maps independently and fully utilizes the information provided by each.
    \item Demonstrates that shallow network features can provide object information, such as object boundary shapes, through simple clustering.
    \item A SIP mechanism based on shallow-layer feature clustering is designed to generate activation maps, producing highly accurate and comprehensive class activation maps.
\end{itemize}

The rest of the paper is organised as follows: The related work is discussed in Sec.~\ref{sec:Related}. We present the proposed method in Sec.~\ref{sec:RegionCAMmethod} and the performance evaluation in Sec.~\ref{sec:Experiments} followed by conclusions in Sec.~\ref{sec:conclusions}.

\section{Related Work}
\label{sec:Related}
Our work focuses on generating fine-grained activation maps for weakly supervised tasks. Therefore, the current class activation methods, WSSS and WSOL will be discussed in this section. 

\subsection{Class Activation Mapping}

The original CAM method was first introduced by \cite{zhou2016learning} for visual explanation of CNNs and object localization through weighting feature maps with classifier parameters. However, it is structure sensitive and it cannot be directly applied to networks without global average pooling layer. As a generalization of the CAM method, some gradient-based \cite{selvaraju2017grad,chattopadhay2018grad,rebuffi2020there,jiang2021layercam} have been proposed. Similar to the CAM \cite{zhou2016learning}, the gradient-based method also weights the features to obtain activation maps \cite{selvaraju2017grad,chattopadhay2018grad} but these methods extract gradients as the weighting coefficient, which eliminates the structural requirements of the CAM. However, most methods generate class activation maps from the last convolutional layer feature maps \cite{zhou2016learning,wang2020score,bany2021eigen,ramaswamy2020ablation,selvaraju2017grad,chattopadhay2018grad,zhang2021group}. These feature maps have low resolution and limited perception of details. Consequently, the produced activation maps lack accurate and complete object information, which constrains their effectiveness in weakly supervised tasks, particularly WSSS that requires precise pixel-level labels \cite{shi2021zoom,rebuffi2020there}. Moreover, the above-mentioned methods are designed primarily for visual explanation, with the objective of faithfully capturing the image regions that contribute to a model's predictions. Therefore, they may incorrectly activate background or other objects instead of highlighting the complete and correct object regions needed for downstream weakly supervised tasks \cite{chen2022score,bany2021eigen,yang2024decomcam}.

There are a few CAM methods specifically designed for weakly supervised tasks. Zoom-CAM \cite{shi2021zoom} utilizes a scalar weighting to generate activation maps in intermediate layer. Similarly, LayerCAM \cite{jiang2021layercam} locate more fine-grained detail object information by fusing multiple activation maps from shallow layers. However, the activation maps from shallow layers contain noise, which may lead to inaccurate activation  \cite{wang2020score,ramaswamy2020ablation,jiang2021layercam}. In addition, Yasuki and Taki \cite{yasuki2024cam} applied the original CAM method into a large-core CNN with data augmentation training strategy to obtain more comprehensive activation maps. Nevertheless, their method requires data augmentation to be applied during network training, which makes it difficult to combine their approach with other existing weakly supervised methods, as most of them are highly customized. Furthermore, these methods still struggle to generate accurate pixel-level activation maps. Specifically, the activation regions are incomplete for large objects, the background is over-activated for small objects, and the boundaries of the activation regions are not aligned with the object boundaries. 

Compared to these methods, our proposed Region-CAM method averages semantic information based on feature clustering results. This not only activates more target regions but also suppresses background regions, thus generating accurate pixel-level activation maps for weakly supervised tasks.

\begin{figure*}[t]
  \centering
   \includegraphics[width=0.95\linewidth]{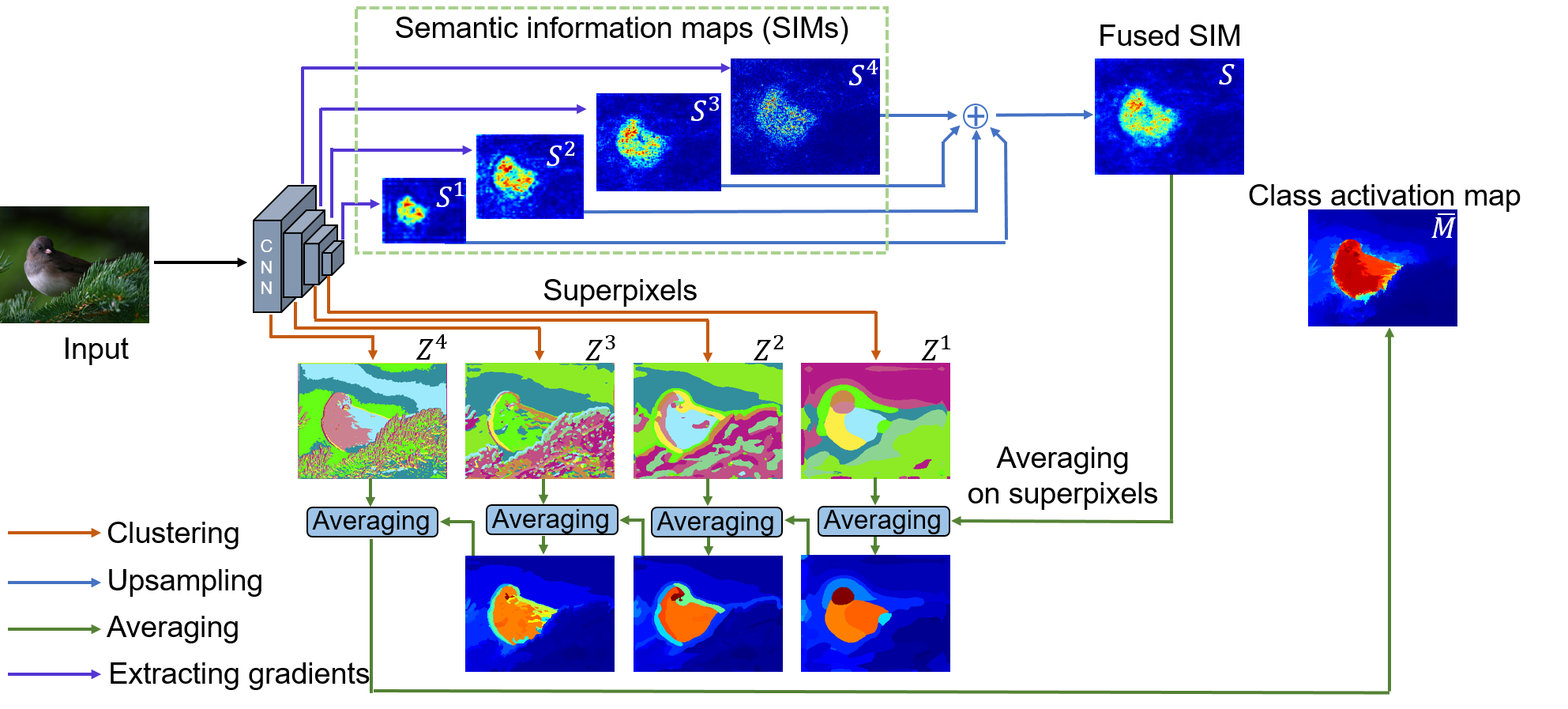}
   \caption{The overall framework of our Region-CAM. Superpixels are firstly obtained from different layers during the forward pass. The SIMs are extracted based on the network gradient, and then fused to generate interim class activation map. Finally, interim class activation map is averaged over different superpixels in turn, to obtain the final class activation map.}
   \label{framework}
\end{figure*}

\subsection{Weakly Supervised tasks}
Image-level labels driven WSSS and WSOL have attracted widespread attention due to their low annotation cost. The ability of CAM methods to generate high-quality activation maps is crucial to the success of these tasks \cite{yasuki2024cam}, as class activation maps readily convert into segmentation seeds and target locations through simple threshold comparisons \cite{wu2021embedded,meng2021foreground}. In the WSSS domain, the segmentation seeds are converted into pseudo pixel-level annotations for subsequent network training to achieve the semantic segmentation task \cite{chen2022self,chang2020weakly,wang2020self,wu2021embedded,lee2021anti,kweon2021unlocking}. However, many WSSS methods \cite{wu2025prompt,wu2024masked,shao2024knowledge,wang2023treating,zhang2021complementary,chen2022class,yang2024foundation,xu2025weakly} still suffer from the limitation that the original CAM method not accurately highlighting the predicted target class pixels. Therefore, this issue must also be taken into account when designing WSSS algorithms. Tong \textit{et al.} \cite{wu2021embedded} explored more comprehensive class-specific activation maps, by an Embedded Discriminative Attention Mechanism (EDAM), which directly integrates the CAM into WSSS network. Hyeokjun \textit{et al.} \cite{kweon2021unlocking} proposed a class-specific adversarial erasing framework, which fully exploits the potential of ordinary classifiers and further explores areas with weaker discriminative power while suppressing the mutual intrusion of activation maps between different categories. Last year, Yang and Gong \cite{yang2024foundation} introduced the Segment Anything Model (SAM) into WSSS to refine segmentation results. However, SAM is trained on the massive SA-1B dataset, which contains over 1 billion masks derived from 11 million images. It is not appropriate to combine SAM with WSSS methods that rely solely on image-level supervision. In addition, Yang and Gong \cite{yang2024foundation} compared their method with fully supervised semantic segmentation approaches. Recently, Zhu \textit{et al.} \cite{zhu2025weakclip} applied the Contrastive Language–Image Pretraining (CLIP) foundation model to refine the activation maps generated by the original CAM method. This foundation model is trained on large-scale image–text pair annotations without pixel-level labels.

Similarity, most works in WSOL have focused on enabling the network to activate more complete object regions based on the original CAM method by designing specialized network architectures and training strategies \cite{yasuki2024cam,chen2024adaptive,xu2022cream,kumar2017hide,xue2019danet,xie2021online,meng2021foreground}. Singh \textit{et al.} \cite{kumar2017hide} forced the network to recognize more object region by hiding the patches where the object is most represented based on the CAM method. Xu \textit{et al.} \cite{xu2022cream} proposed Class Reactivation Mapping (CREAM) to improve the activation ability of the CAM method and activate more object regions. Chen \textit{et al.} \cite{chen2024adaptive} combined the Adversarial Learning Mechanism (ALM) and the Targeted Learning Mechanism (OLM) to achieve coarse-to-fine optimization of CAM results through a triple adaptive region mechanism. 

To achieve better segmentation and localization results, many weakly supervised methods explicitly consider the limitation of the original CAM method when designing their networks. Therefore, we believe that developing an activation algorithm capable of activating a greater proportion of target regions cannot only fundamentally address the insufficient activation problem of the original CAM method and improve the performance of existing weakly supervised algorithms, but also provide valuable contributions to future research in weakly supervised learning.

\section{Methodology}
\label{sec:RegionCAMmethod}
In this section, we first revisit the widely used original CAM method and the related activation method Grad-CAM. We then introduce our proposed method, explaining how we obtain the SIMs in Sec.~\ref{Semanticinformationmap} and how we enhance target class activations and minimize false positive activations through our designed SIP mechanism in Sec.~\ref{Semanticinformationpropagation}. The pipeline of our proposed method is shown in Figure~\ref{framework}.

\subsection{Revisit CAM and Grad-CAM}
\label{RevisitCAMandGrad-CAM}

Let $I$ represent an input image and $F_l \in \mathbb{R}^{ h\times w \times k}$ represent the feature maps extracted from layer $l$ of a convolutional classification network. The prediction score for the target category $c$, prior to applying the softmax function, is given by $y^c$. The $W^{c}$ is the classification weight corresponding to category $c$ with channel $k$. The activation map $M^c$ of category $c$ generated by the original CAM method can calculated as: 
\begin{equation}
	M^c = ReLU\left(\sum_k W^{c} F_{last}\right),
	\label{eq:CAMmehtod}
\end{equation}
where, $F_{last}$ is extracted from the last convolutional layer of the network. Other methods, such as Grad-CAM~\cite{selvaraju2017grad}, Grad-CAM++~\cite{chattopadhay2018grad}, and LayerCAM~\cite{jiang2021layercam}, also generate activation maps based on the principle of weighting feature maps. However, these methods utilize gradients instead of classification parameters and process gradients differently. The gradient of the score $y^c$ with respect to the feature map $F_l$ at position $(i,j)$ can be expressed as:
\begin{equation}
	g_{ij}^{c} =  \frac{\partial y^c}{\partial F_{(ij)}^l}.
	\label{eq:grad}
\end{equation}
Consider Grad-CAM as an illustrative example, the channel-wise weight $\alpha^c \in \mathbb{R}^{ 1\times 1 \times k}$ for the feature map $F_l$ can be calculated by averaging the gradients of all locations in the corresponding activation maps:
\begin{equation}
  \alpha^c = \frac{1}{N}\sum_i\sum_j g_{ij}^{c},
  \label{eq:weightalpha}
\end{equation}
where $N$ denotes the number of spatial points on one channel of the feature map $F_l$. Finally, the class activation map of category $c$ can be expressed as:
\begin{equation}
  M^c = ReLU\left(\sum_k \alpha_k^c \cdot F_{l}\right).
  \label{eq:gradcam}
\end{equation}
In Grad-CAM and Grad-CAM++, $F_{l}$ still comes from the last layer of the network to ensure the reliability of the activation map. Differentially, LayerCAM obtains activation maps by fusing pixel-wise weighted feature maps from multiple layers.

However, we argue that features and gradients should be considered separately to maximize their effects, because gradients and features contain different object information and they can complement each other based on our experiments. Specifically, gradients imply semantic information about the target categories, according to~\cite{chattopadhay2018grad}, if there is a spatial feature point $p$ from feature map $F_{l}$ that contributes to an existing object $O_c$ belonging to category $c$, the gradient of point $p$ is expressed as $g_p^c$. A larger gradient $g_p^c$ means that small changes in this point $F^{l}_{p}$ would cause significant changes in the score $y^c$, which also implies that this spatial point $p$ contains important and accurate semantic information about the specific object $O_c$. We believe that the value of the gradient $g_{p}^{c}$ is an indication of the strength of the semantic information.

Moreover, we found that clustering features from shallow networks can generate superpixel regions that provide prior boundary information. Such boundary information is object-independent and remains consistent across classes. Therefore, we believe that when generating activation maps for specific objects, gradients and network features should be processed separately to produce more complete activation maps by complementing their information.

\subsection{Semantic Information Maps}
\label{Semanticinformationmap}
\begin{figure*}[tb]
    \centering
    \begin{minipage}{0.089\textwidth} 
            \centering
            \includegraphics[width=\textwidth]{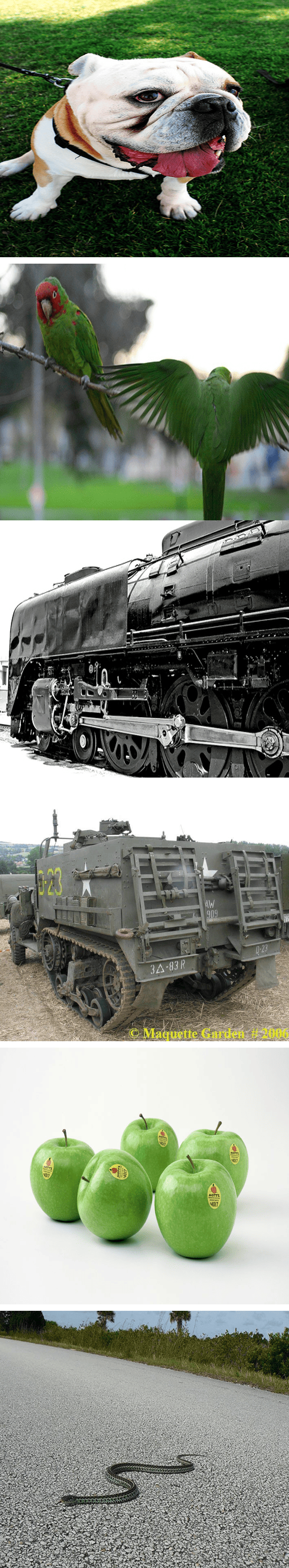}\\            
            {(a)}\\
            {Input}
    \end{minipage}
    \begin{minipage}{0.45\textwidth}
        \centering
        \includegraphics[width=\linewidth]{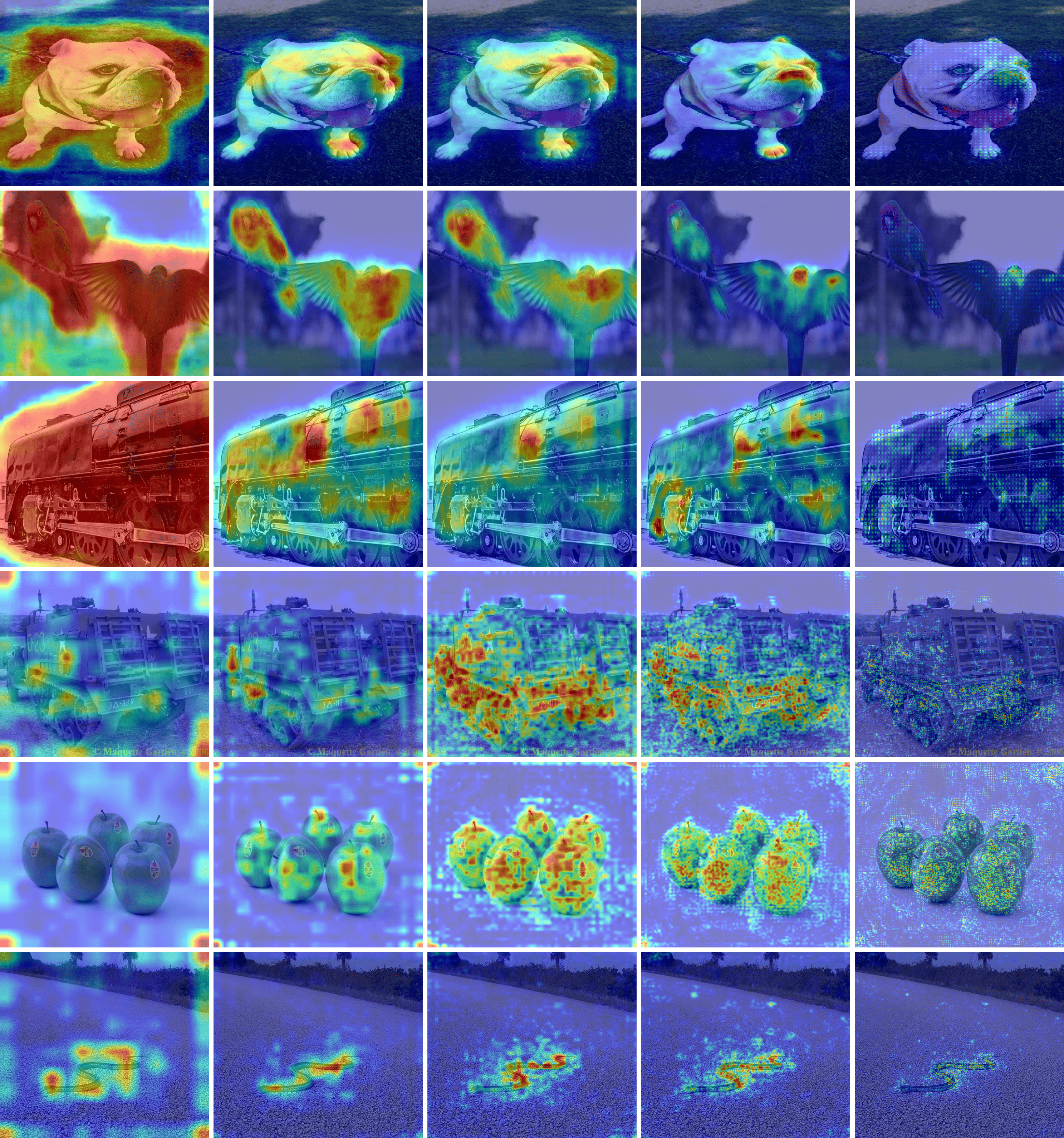}\\ 
        {(b)\hspace{0.14\textwidth} (c)\hspace{0.14\textwidth} (d)\hspace{0.14\textwidth} 
        (e)\hspace{0.14\textwidth} (f)}\\
        {Semantic information maps from different layers}
    \end{minipage}
    \begin{minipage}{0.45\textwidth}
        \centering
        \includegraphics[width=\linewidth]{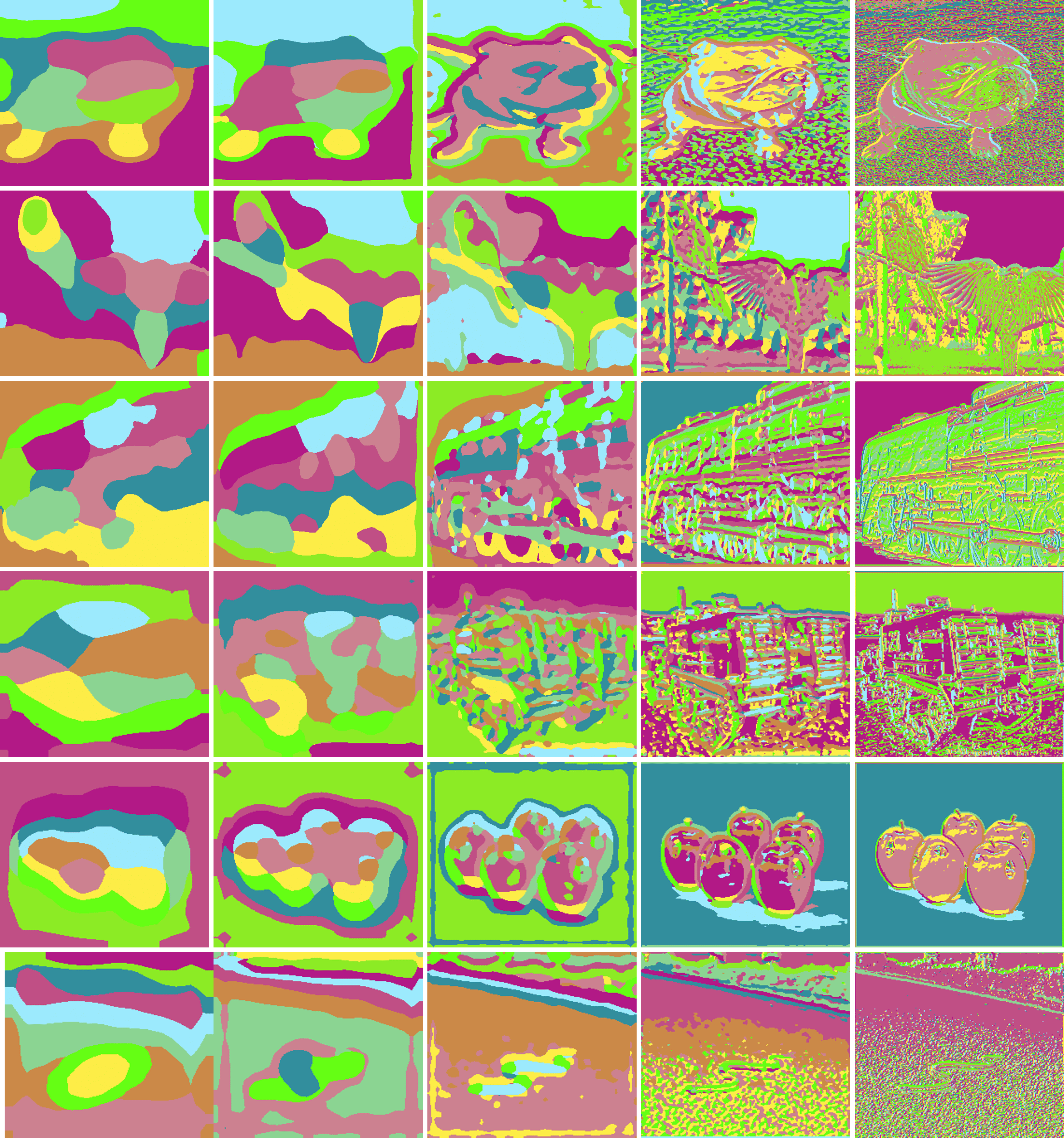}\\ 
        {(g)\hspace{0.14\textwidth} (h)\hspace{0.14\textwidth} (i)\hspace{0.14\textwidth} 
        (j)\hspace{0.14\textwidth}(k)}\\
        {Clustering results from different layers}
    \end{minipage}
    \caption{\textbf{Visualization of SIMs and clustering results form different layers. (a) Input image; Columns (b)-(f): The SIMs are generated from the deep to the shallow layers - block\_5, block\_4, block\_3, block\_2, and block\_1 respectively. Spatial points highlighted in red denote stronger target semantic information, whereas those in blue denote little or no target semantic information; Columns (g)-(k): The clustering results are also generated from the deep to the shallow layers - block\_5, block\_4, block\_3, block\_2, and block\_1 respectively. In both the SIMs and clustering results, the images in the top three rows are generated by ResNet38 model and the input images from the PASCAL VOC 2012 \cite{everingham2010pascal} \cite{wu2019wider}, while those in the bottom three rows are generated by VGG16 based on the ILSVRC2012 \cite{russakovsky2015imagenet} \cite{simonyan2014very}.}}
    \label{semanticClustering}
\end{figure*}

Based on the above analysis, we propose our novel class activation mapping method, Region-CAM. Our method is divided into two parts: SIM acquisition, and SIP execution. The gradients of category $c$ with respect to the feature map $F_l$ is obtained through Eq. (\ref{eq:grad}). According to \cite{chattopadhay2018grad}, and \cite{jiang2021layercam}, positive gradient values represent the semantic information that has a positive impact on the predicted. Therefore, by summing the non-negative gradients across channels, the SIM $S^c_l$, which encapsulates the semantic information of the target category $c$ from $l$ layer, can be calculated as:
\begin{equation}
  S^{c}_{l} = \sum_k^K ReLU(\frac{\partial y^c}{\partial F_{k}}),
  \label{eq:semanticmaps}
\end{equation}
where $K$ represents the number of channels in the feature map $F_{l}$. Columns (b)-(f) in Fig.~\ref{semanticClustering} illustrate SIMs across different layers, arranged from deep to shallow, left to right. Spatial points highlighted in red indicate stronger target semantic information. In contrast, spatial points highlighted in blue correspond to points with little to no target semantic information. The map from the deep layer effectively identifies object locations, while the shallow layer map captures finer object details due to higher spatial resolution. Consequently, a complete and fine-grained detailed SIM $S^{c}$ can be obtained by fusing SIMs from different layers. The fusion process is expressed as:
\begin{equation}
  S^c = \sum_l S^{c}_{l}.
  \label{eq:OriginlaCAM1}
\end{equation}

However, we also observe from the bottom three rows images of Fig. \ref{semanticClustering} (in columns (b)-(f)) that the SIMs are scattered and they contain more noise compared to those top three rows. The possible reason is that these three layers are produced by a network with max pooling layers \cite{zhou2016learning}. These noise affect the activation results and object localization \cite{ramaswamy2020ablation,wang2020score,jiang2021layercam}, and further affect the generation of accurate pixel-level activation maps to provide good segmentation seeds for WSSS. To address the above two problems, we design a SIP mechanism based on network features to denoise the fused SIM, $S^c$, and propagate the semantic information to the object region based on the superpixel regions.

\subsection{Semantic Information Propagation}
\label{Semanticinformationpropagation}
In order to denoise the $S^c$ and enhance the activation of the target area, the SIP mechanism is performed based on superpixel regions via a simple averaging operation. The superpixel regions are created through unsupervised clustering of network features utilizing the K-means algorithm. Given a set of spatial feature points $\{p_1,p_2,...,p_q \}$, where each point $p_i \in \mathbb{R}^d$, the objective of K-means is to partition the points into $m$ clusters, denoted $Z_1, Z_2,...Z_m$. The objective can be expressed as:
\begin{equation}
	J = argmin\left[\sum_j^m\sum_{p_i \in Z_j} \left( 1 - \frac{p_i \cdot \mathbf{\mu}_j}{\|p_i\| \|\mathbf{\mu}_j\|} \right) \right],
	\label{eq:kmeans}
\end{equation}
where $\mu_j$ is centroid of cluster $Z_j$, $p_i$ is the spatial point from feature map $F$ at a certain layer, and $J$ is the total intra-cluster variance. The $\mu_j$ can be calculated as:
\begin{equation}
	\mu_j = \frac{1}{N_j}\sum_{p_i \in Z_j} p_i, 
	\label{eq:mu}
\end{equation}
where $N_j$ is the number of points in cluster $j$. Since the K-means algorithm groups points into clusters by minimizing the distance within the cluster, after clustering the spatial points within the same cluster contain similar semantic information~\cite{achanta2012slic}.

Moreover, we observe that the clustering results gradually reveal the details of the object from deep to shallow layers. Columns (g) to (k) in Figure \ref{semanticClustering} show the clustering results of feature maps from different layers. The clustering results are generated from the deep to the shallow layers from left to right. In the deepest layer, the clustering results clearly show that the spatial points in the object and background area belong to two different cluster centroids. Executing the SIP mechanism on the clustering results of the deep network ensures that the semantic information is propagated in the target-related areas, although the superpixel boundaries are not aligned with the boundaries of the object. The superpixels $Z_j$ from the shallowest layer show clear object detail information, specifically, the boundaries of superpixels align with object boundaries and region sizes, as shown in the column (k) of Figure ~\ref{semanticClustering}. Aligned superpixel boundaries further ensure that semantic information propagates along precise boundaries, resulting in pixel-level accurate activation maps. Therefore, we first perform SIP mechanism on deep layer superpixels $Z_j$ to concentrate the semantic information, and filter out the noise in the process, followed by gradually averaging the SIM in shallower layer $Z_j$ for further refinement. The process of SIP of the $S^c$ in the layer $l$ superpixels is as follows: 
\begin{equation}
\forall j \in \{1,...,m\}:  \bar{S^c} = \frac{1}{N_j}\sum_{i\in Z^l_j} S^c_i,
  \label{eq:RegionCAM}
\end{equation}
where $Z^l_j$ is the superpixel $j$ obtained by clustering features from the layer $l$ of the network and $i$ is a pixel in the superpixel cluster $Z^l_j$. We average the $\bar{S^c}$ over a series of superpixels from different layers to obtain the class activation map $\bar{M}$.

\section{Performance Evaluation}
\label{sec:Experiments}

In this section, we verify the performance of Region-CAM in generating accurate pixel-level activation maps by conducting a segmentation seeds generation experiment presented in Sec.\ref{Experiments1-1}. In addition, we verify the applicability of accurate object maps in another weakly supervised learning application, namely weekly supervised object localization as presented in Sec.\ref{Experiments1-2}. Finally, we evaluate the ability of the proposed Region-CAM method to accurately identify discriminative regions from the perspective of classification by conducting an object occlusion experiment as presented in Sec.~\ref{Experiments1-3}.

\subsection{Generating Segmentation Seeds}
\label{Experiments1-1}

Generating semantic segmentation seeds is the most important step in WSSS, and it is also where WSSS research reflects its algorithmic innovation. Therefore, to verify the performance of our algorithm in generating accurate pixel-level class activation maps, we also follow the WSSS setting to generate semantic segmentation seeds \cite{chen2022self,ahn2019weakly,zhao2024sfc,kweon2023weakly,chen2024adaptive,xu2022multi}.

The PASCAL VOC 2012 \cite{everingham2010pascal} and MS COCO 2014~\cite{lin2014microsoft} datasets been used to evaluate the performance of generating segmentation seeds. These two datasets are benchmarks in the semantic segmentation field, and they have 21 and 81 categories with respectively. The experiments were conducted based on baseline ResNet-38 model~\cite{wu2019wider}, which is a pure classification network trained on both datasets without WSSS-specific optimization. The model weights can be downloaded through \href{https://github.com/halbielee/EPS}{this link}\footnote{\href{https://github.com/halbielee/EPS}{https://github.com/halbielee/EPS}} \cite{lee2021railroad}. For implementation details, we keep the original size of the image and transform them to [0,1], then normalize them with mean vector [0.485, 0.456, 0.406] and standard deviation vector [0.229, 0.224, 0.225]. The Resnet38 model contains 5 blocks, we obtain SIMs from \textit{block\_2}, \textit{block\_3}, \textit{block\_4} and \textit{block\_5} respectively, and fuse them as the SIM for target categories. We perform $tanh$ function to scale the first three SIMs before fusing them. We extract features from \textit{block\_1}, \textit{block\_2}, \textit{block\_3} and \textit{block\_4} and cluster these features with 10 clustering centroids to obtain superpixel regions respectively. Finally, a class activation map of background is manually set and concatenated with the target activation maps. The segmentation seeds mask is obtained by executing the $argmax$ function on the concatenated class activation map, which is a common practice in WSSS domain \cite{chen2022self,wang2020self,chen2024adaptive}.

\begin{table}
\centering
    \begin{minipage}[t]{0.48\textwidth}
    \centering 
    \caption{Comparison with other activation methods under the same model on mIoU (\%). These results are the best that can be achieved with these methods, based on common WSSS settings. * Indicates result is from the original paper.}
	\begin{tabular}{lcccc}
        \hline
	\multirow{2}{*}{Method} & \multirow{2}{*}{Backbone} & \multicolumn{2}{c}{VOC} & COCO\\
	\cline{3-5}
		& & Train& Val.& Val.\\
		\hline
		CAM  & ResNet-38 & 46.51 & 45.30 & 20.15\\
		Grad-CAM  & ResNet-38 & 44.83 & 42.83 & 26.17\\
		Grad-CAM++  & ResNet-38 & 43.61 & 41.43 & 25.54\\
         Zoom-CAM\cite{shi2021zoom}* & VGG16 & 49.0 & - & - \\
		LayerCAM & ResNet-38 & 54.02 & 52.34 & 31.43\\
	\hline
		Region-CAM & ResNet-38 & \textbf{60.12} & \textbf{58.43} & \textbf{36.38}\\
	\hline
	\end{tabular}
	\label{tab:1}
    \end{minipage}
    \hfill
    \begin{minipage}[t]{0.48\textwidth}
    \centering
        \caption{Evaluation (mIoU (\%)) of different segmentation seeds on PASCAL VOC 2012 training set with the State-of-the-art WSSS Methods.}
    \resizebox{\textwidth}{!}{
	\begin{tabular}{lccc}
	\hline
		Method & Pub. & Backbone & mIoU(\%) \\
        \hline
	SCE \cite{chang2020weakly} & CVPR20 & ResNet-38 & 50.9\\
		SEAM \cite{wang2020self} & CVPR20 &ResNet-38 & 55.4\\
        AdvCAM \cite{lee2021anti} & CVPR21 & ResNet-50 & 55.6\\
		ECS \cite{sun2021ecs} & CVPR21 &ResNet-38 & 56.6\\
		SIPE \cite{chen2022self} & CVPR22 &ResNet-50 & 58.6 \\ 
            MCTformer \cite{xu2022multi} & CVPR22 &Transformer & 61.7 \\
		D2CAM \cite{wang2023treating} & CVPR23 &ResNet-50 & 62.3 \\
        SSC \cite{chen2024spatial} &  TIP24 & ResNet-50 &  58.3 \\
		\hline
            Baseline + CAM & & ResNet-38 &46.5 \\
		Baseline + Region-CAM & & ResNet-38 & 60.1\\
		SIPE + Region-CAM & & ResNet-50 &62.8\\
            MCTformer + Region-CAM & & Transformer & 64.2 \\
		\hline
	\end{tabular}}
	\label{tab:2}
    \end{minipage}
\end{table}

\subsubsection{Comparison with Widely Used CAM Methods}

\begin{figure}[h]
    \centering
    \begin{minipage}{\textwidth} 
        \begin{minipage}{0.06\textwidth} 
            \raggedleft
            \rotatebox{90}{\scriptsize \centering Input} 
        \end{minipage}%
        \hspace{0.01em}
        \begin{minipage}{0.88\textwidth} 
            \centering
            \includegraphics[width=\textwidth]{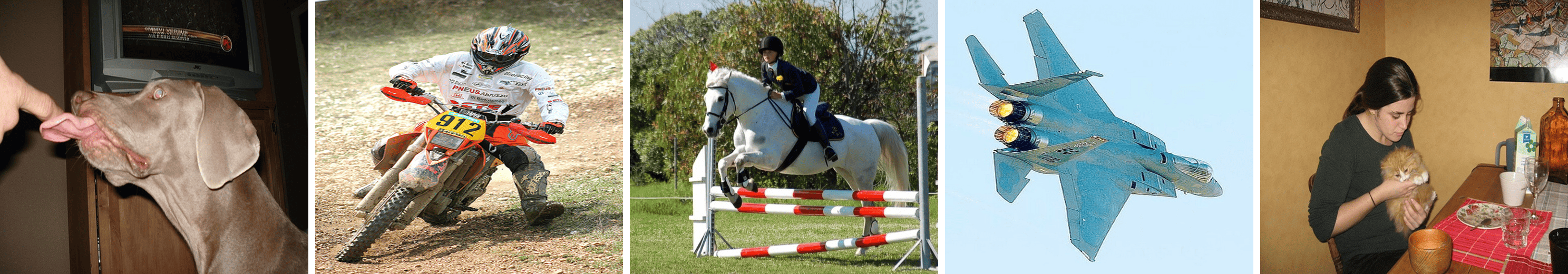} 
        \end{minipage}
    \end{minipage}
    \vspace{0.2em} 
    \begin{minipage}{\textwidth} 
        \begin{minipage}{0.06\textwidth} 
            \raggedleft
            \rotatebox{90}{\scriptsize \centering CAM}
        \end{minipage}%
        \hspace{0.01em}
        \begin{minipage}{0.88\textwidth} 
            \centering
            \includegraphics[width=\textwidth]{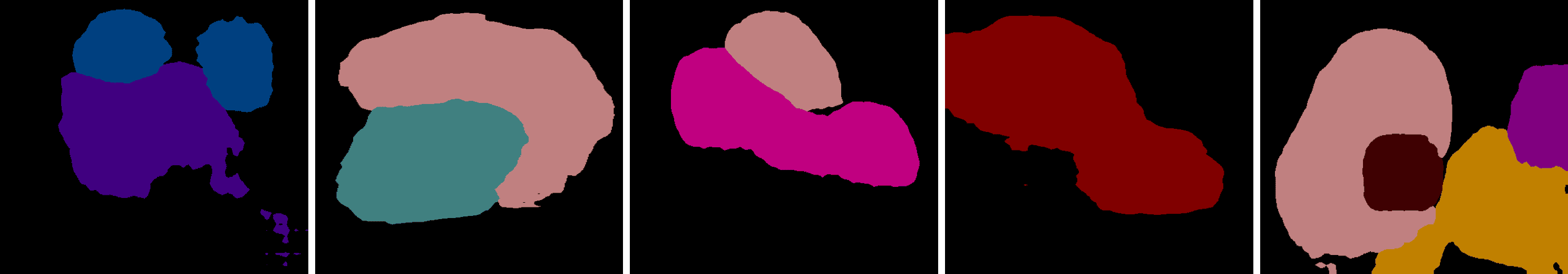}
        \end{minipage}
    \end{minipage}

    \vspace{0.2em} 

    \begin{minipage}{\textwidth} 
        \begin{minipage}{0.06\textwidth} 
            \raggedleft
            \rotatebox{90}{\scriptsize \centering Grad-CAM}
        \end{minipage}%
        \hspace{0.01em}
        \begin{minipage}{0.88\textwidth} 
            \centering
            \includegraphics[width=\textwidth]{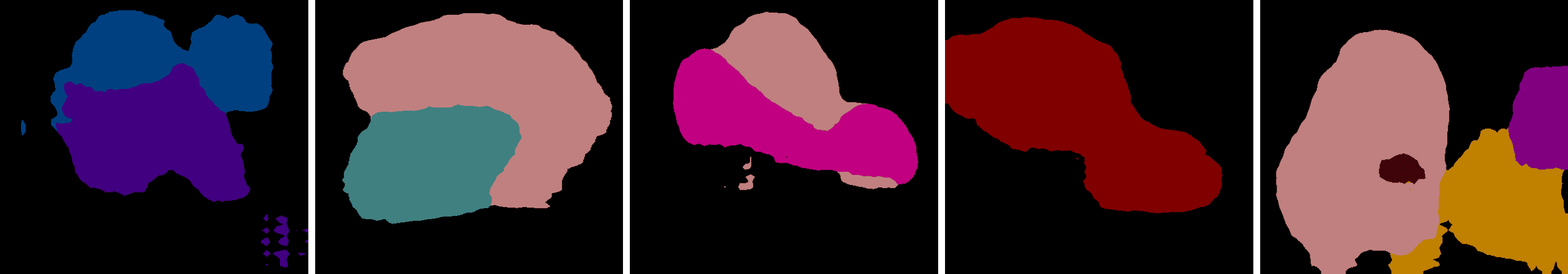}
        \end{minipage}
    \end{minipage}

    \vspace{0.2em} 

    \begin{minipage}{\textwidth} 
        \begin{minipage}{0.06\textwidth} 
            \raggedleft
            \rotatebox{90}{\scriptsize \centering Grad-CAM++}
        \end{minipage}%
        \hspace{0.01em}
        \begin{minipage}{0.88\textwidth} 
            \centering
            \includegraphics[width=\textwidth]{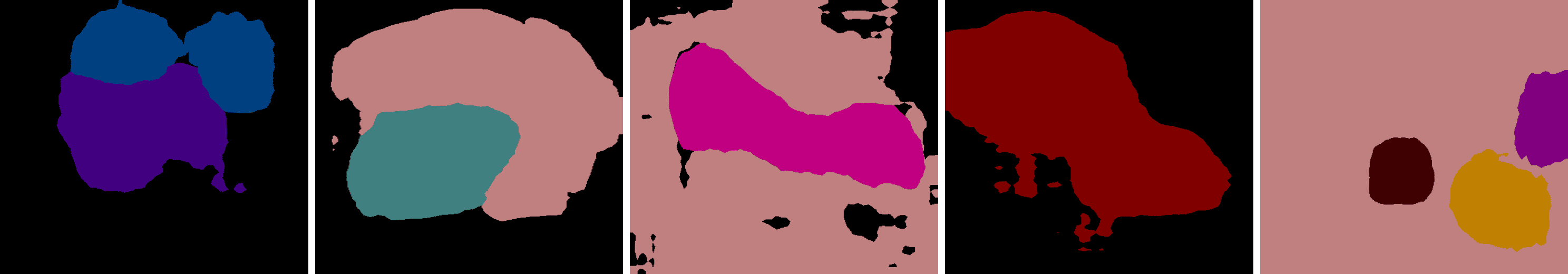}
        \end{minipage}
    \end{minipage}
    
    \vspace{0.2em} 

    \begin{minipage}{\textwidth} 
        \begin{minipage}{0.06\textwidth} 
            \raggedleft
            \rotatebox{90}{\scriptsize \centering LayerCAM}
        \end{minipage}%
        \hspace{0.01em}
        \begin{minipage}{0.88\textwidth}
            \centering
            \includegraphics[width=\textwidth]{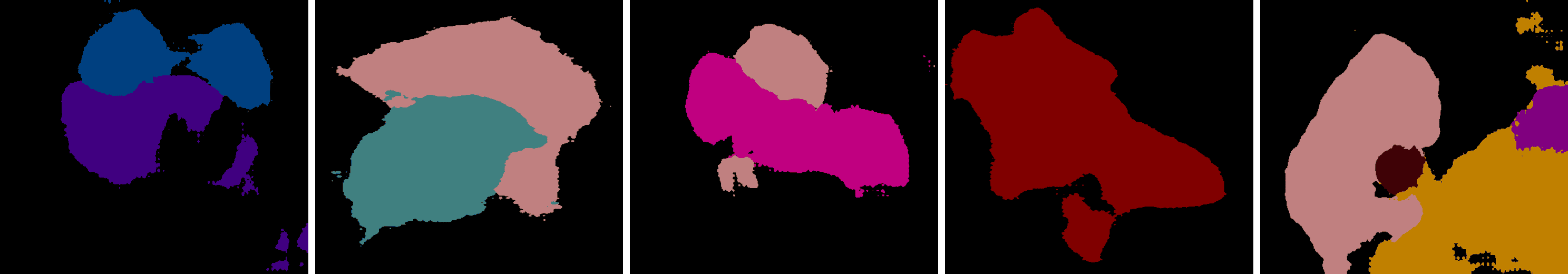}
        \end{minipage}
    \end{minipage}
    
    \vspace{0.2em} 
    
    \begin{minipage}{\textwidth} 
        \begin{minipage}{0.06\textwidth} 
            \raggedleft
            \rotatebox{90}{\scriptsize \centering Region-CAM}
        \end{minipage}%
        \hspace{0.01em}
        \begin{minipage}{0.88\textwidth} 
            \centering
            \includegraphics[width=\textwidth]{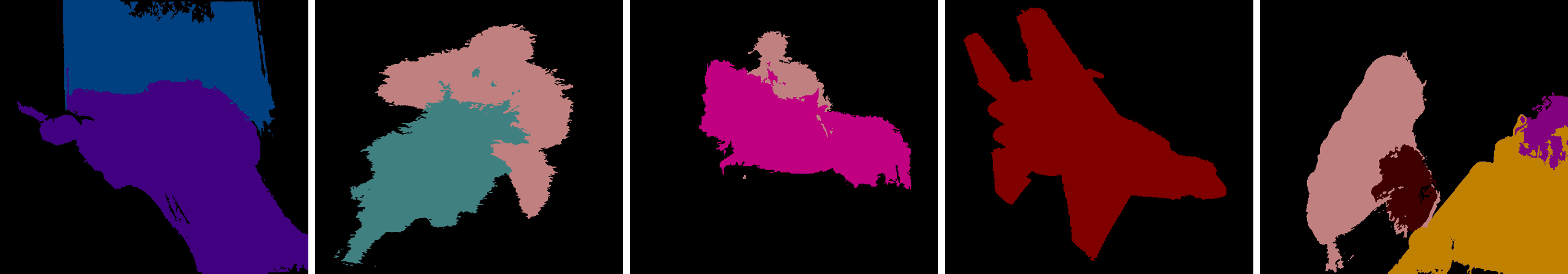}
        \end{minipage}
    \end{minipage}

    \vspace{0.2em} 
    
    \begin{minipage}{\textwidth} 
        \begin{minipage}{0.06\textwidth} 
            \raggedleft
            \rotatebox{90}{\scriptsize GT}
        \end{minipage}%
        \hspace{0.01em}
        \begin{minipage}{0.88\textwidth} 
            \centering
            \includegraphics[width=\textwidth]{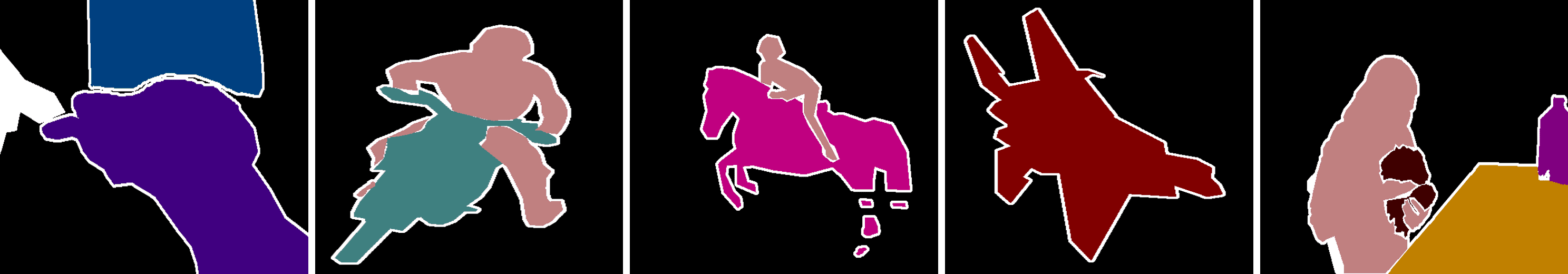}
        \end{minipage}
    \end{minipage}
    
    \caption{Qualitative comparison for pseudo segmentation labels on PASCAL 2012.}
    \label{iniMandrefM}
\end{figure}

We evaluate the performance of our proposed Region-CAM and compare with other widely used activation methods, including CAM \cite{zhou2016learning}, Grad-CAM \cite{selvaraju2017grad}, Grad-CAM++ \cite{chattopadhay2018grad}, Zoom-CAM \cite{shi2021zoom}, and LayerCAM \cite{jiang2021layercam}, in terms of the quality of generated segmentation seeds \cite{chen2022self}. As shown in Table~\ref{tab:1}, Region-CAM achieved 60.12\%, 58.43\%, and 36.38\% mIoU on the PASCAL VOC train set, validation set and MS COCO validation set, respectively. It outperforms CAM by 13.61\%, 13.13\% and 16.23\% on the three sets, respectively. Region-CAM is also 6.1\%, 6.09\% and 4.95\% better than LayerCAM. This clearly demonstrates the superiority of Region-CAM in producing fine-grained activation maps. Figure~\ref{iniMandrefM} illustrates the results of segmentation seeds produced by each activation method. The outputs from CAM, Grad-CAM, and Grad-CAM++ lack object details and mislabel background regions as part of the object, resulting in lower accuracy. Although LayerCAM has fewer mislabelings, the results still lack the level of detail achieved with Region-CAM. The segmentation seeds generated using Region-CAM closely resemble the Ground-Truth (GT) labels, with object edges that are sharp and well-aligned with the actual GT boundaries. Table~\ref{tab:eachmiou} shows the IoU results of each class generated by each activation method on the VOC training set. As shown in this table, Region-CAM enhances activation performance across all categories, with the most pronounced improvements observed for “plant” and “sheep” classes. This phenomenon can be attributed to the greater distinctiveness of features in these categories, which facilitates the formation of consistent and spatially continuous superpixel regions after clustering.

\begin{table*}[tb]
	\centering
    \caption{Comparison with other activation methods based on ResNet-38 model on PASCAL VOC 2012 training set with mIoU (\%) evaluation metrics. These results are the best that can be achieved with these methods, which is common settings on WSSS.}
        \setlength{\tabcolsep}{1.3pt}
        \renewcommand{\arraystretch}{1.5}
    \resizebox{\textwidth}{!}{
	\begin{tabular}{l|c|c|c|c|c|c|c|c|c|c|c|c|c|c|c|c|c|c|c|c|c|c}
        \hline
	Method  & \multicolumn{1}{c}{\rotatebox{60}{bkg}} & \multicolumn{1}{c}{\rotatebox{60}{aero}} & \multicolumn{1}{c}{\rotatebox{60}{bike}} & \multicolumn{1}{c}{\rotatebox{60}{bird}} & \multicolumn{1}{c}{\rotatebox{60}{boat}} & \multicolumn{1}{c}{\rotatebox{60}{bottle}} & \multicolumn{1}{c}{\rotatebox{60}{bus}} & \multicolumn{1}{c}{\rotatebox{60}{car}} & \multicolumn{1}{c}{\rotatebox{60}{cat}} & \multicolumn{1}{c}{\rotatebox{60}{chair}} & \multicolumn{1}{c}{\rotatebox{60}{cow}} & \multicolumn{1}{c}{\rotatebox{60}{table}} & \multicolumn{1}{c}{\rotatebox{60}{dog}} & \multicolumn{1}{c}{\rotatebox{60}{horse}} & \multicolumn{1}{c}{\rotatebox{60}{mbike}} & \multicolumn{1}{c}{\rotatebox{60}{person}} & \multicolumn{1}{c}{\rotatebox{60}{plant}} & \multicolumn{1}{c}{\rotatebox{60}{sheep}} & \multicolumn{1}{c}{\rotatebox{60}{sofa}} & \multicolumn{1}{c}{\rotatebox{60}{train}} & \rotatebox{60}{tv} & mIoU \\
        \hline
		CAM  & 73.36 & 35.49 & 26.62 & 38.19 & 28.96 & 46.83 & 64.41 & 50.41 & 53.35 & 24.10 & 50.53 & 44.52 & 55.02& 49.85 & 59.15 & 47.08 & 37.91 & 55.25 & 44.03 & 52.35 & 39.27 & 46.51 \\
            Grad-CAM & 73.03 & 37.58 & 28.03 & 40.31 & 31.0 &   42.59 & 61.16 & 45.19 & 45.35 & 23.10 & 51.07 & 39.01 & 53.28 & 47.69 & 59.32 & 41.10 & 35.47 & 55.91 & 37.19 & 54.87 & 38.69 & 44.83 \\
	    Grad-CAM++ & 66.15 & 39.85 & 25.04 & 40.40 &      31.24 & 34.92 & 60.21 & 39.97 & 44.43 & 23.81 & 51.48 & 42.13 & 47.75 & 47.69 & 60.70 & 29.13 & 31.27 & 57.63 & 43.59 & 54.55 & 43.91 & 43.61\\
		LayerCAM & 78.89 & 46.82 & 31.33 & 51.21 & 37.37 & 52.36 & 70.71 & 56.48 & 62.30 & 26.99 & 65.94 & 47.42 & 65.23 & 62.63 & 68.29 & 47.40 & 34.93 & 75.20 & 45.32 & 58.53 & 40.11 & 54.02 \\
		Region-CAM &  \textbf{80.60} &  \textbf{55.73} &  \textbf{33.57} &  \textbf{62.71} &  \textbf{43.43} &  \textbf{60.25} &  \textbf{75.44} &  \textbf{59.57} &  \textbf{70.52} &  \textbf{31.25} &  \textbf{73.37} &  \textbf{49.23} &  \textbf{75.73} &  \textbf{70.35} &  \textbf{72.74} &  \textbf{52.64} &  \textbf{44.43} &  \textbf{81.07} &  \textbf{52.85} &  \textbf{62.15} &  \textbf{51.96} &  \textbf{60.12} \\
	\hline
	\end{tabular}}
	\label{tab:eachmiou}
\end{table*}

\begin{figure*}[h]
    \centering
    \begin{minipage}{0.44\textwidth}
        \centering
        \includegraphics[width=\linewidth]{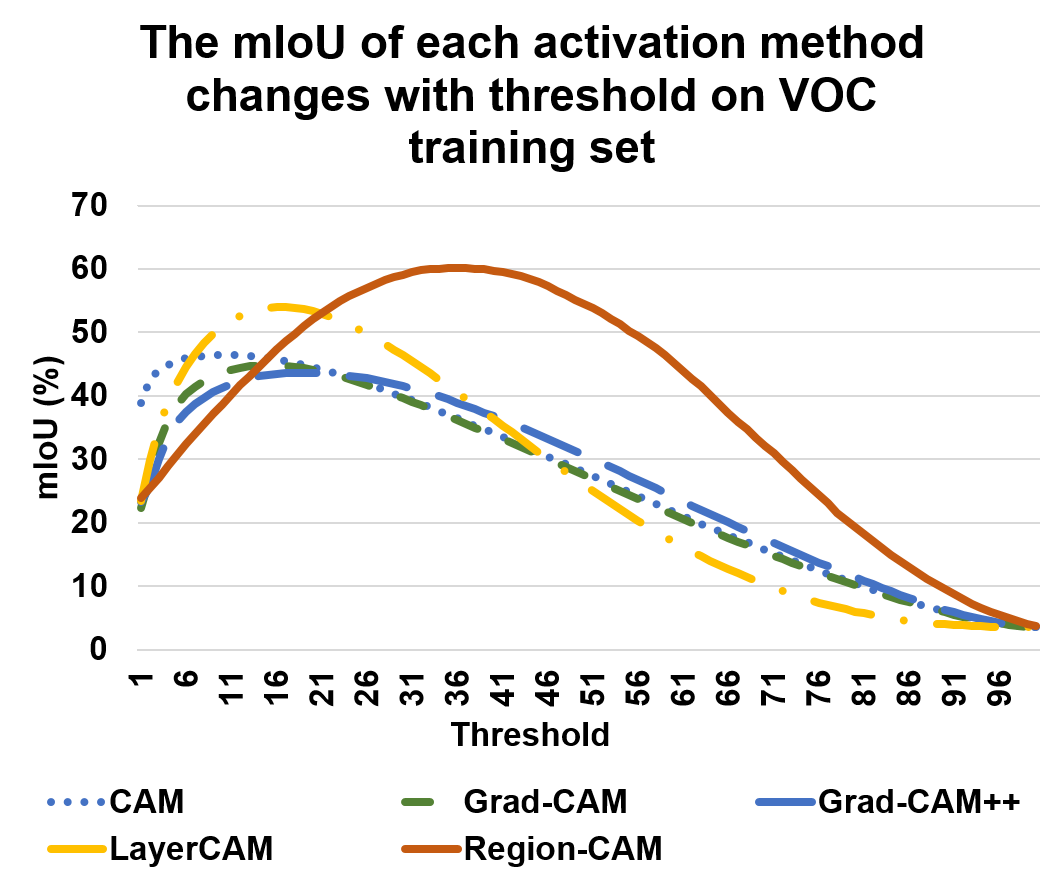}\\ (a) The mIoU changes with threshold on VOC training set 
    \end{minipage}
    \begin{minipage}{0.44\textwidth}
        \centering
        \includegraphics[width=\linewidth]{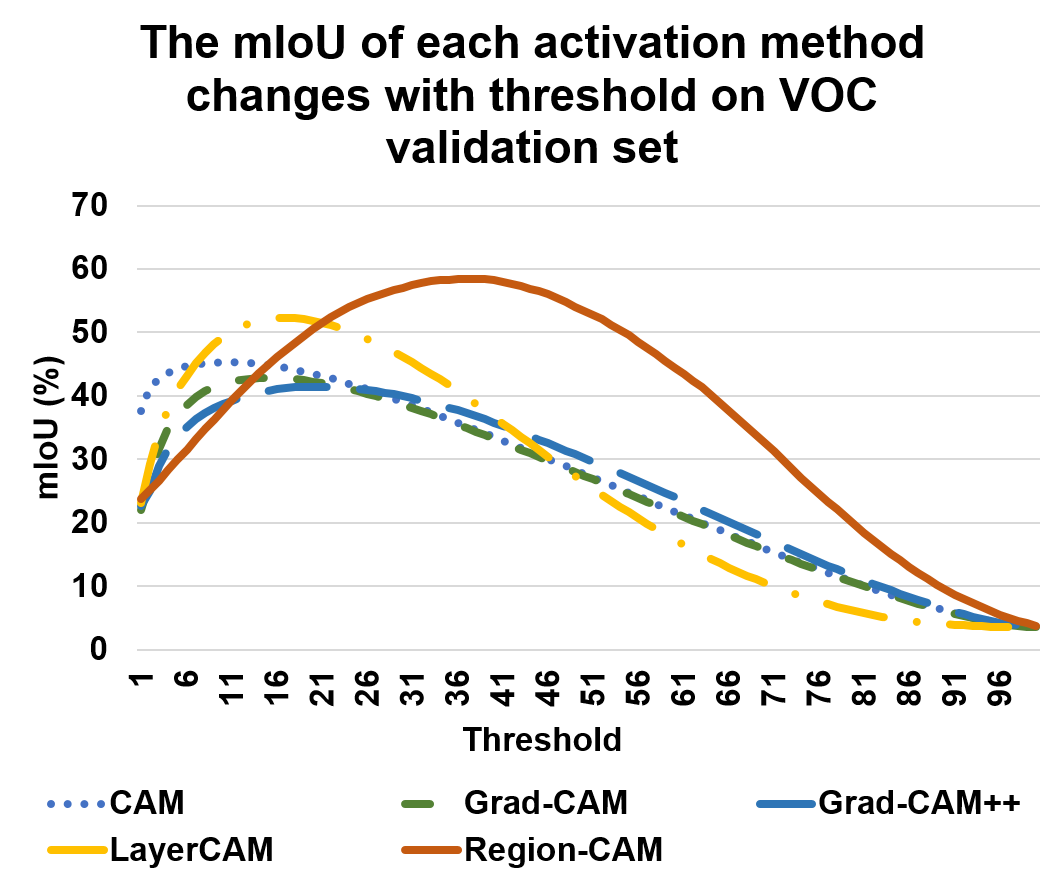}\\ (b) The mIoU changes with threshold on VOC validation set
    \end{minipage}
    \caption{\textbf{The mIoU of each activation method changes with threshold on VOC dataset.}}
    \label{mIoUchangewithThreshold}
\end{figure*}

According to the common practice in the WSSS field \cite{wang2020self,meng2019weakly,wang2017weakly,cai2023ssdb}, manual background thresholding is typically applied when converting activation maps into segmentation seeds. In previous studies, the best-performing case is reported. However, the choice of threshold has a impact on the evaluation, since the mIoU values vary considerably across different threshold settings. Reliance on a fixed threshold may introduce bias and hinder a fair assessment of the robustness of different methods. To ensure a more comprehensive evaluation, we report the mIoU of each activation algorithm across a range of thresholds, as illustrated in Figure \ref{mIoUchangewithThreshold}. This presentation allows a clearer understanding of how sensitive each algorithm is to threshold selection and offers a more objective comparison of their overall performance. 

As illustrated in Figure~\ref{mIoUchangewithThreshold} and Table~\ref{TheaveragevalueofmIoUinthethreshold}, although LayerCAM achieves better peak performance than CAM, Grad-CAM, and Grad-CAM++, its average mIoU remains similar to these methods. This phenomenon arises because the activation maps aggregated by LayerCAM exhibit limited consistency; different layers exhibit varying activation levels across spatial points. After aggregation, only a small subset of spatial points retains high activation values, whereas the majority remain weakly activated. Consequently, as the background threshold increases, the mIoU of the derived semantic seeds progressively declines, and at higher threshold settings, the performance of LayerCAM is inferior to that of CAM, Grad-CAM, and Grad-CAM++. In contrast, Region-CAM demonstrates superior overall performance, as its SIP mechanism enables the spread of activation values across superpixel regions, resulting in more complete region activation. A detailed analysis of Region-CAM is presented in Sec.~\ref{AblationStudy}. The next section will highlight how Region-CAM enhances the performance of existing state-of-the-art WSSS methods.

\begin{table}[t]
	\centering
    \caption{Comparison of the accuracy of the segmentation seeds generated by the $S$ and the $\bar{M}$ on mIoU (\%)}
	\begin{tabular}{llccc}
		\hline
		\multicolumn{2}{c}{Dataset} & CAM & $S$ & $\bar{M}$\\
		\hline
		\multirow{2}{*}{VOC} & Train & 46.51 & 54.09 & \textbf{60.12}\\
		& Val. & 45.30 & 52.78 & \textbf{58.43}\\
		\hline
		COCO & Val. & 20.15 & 32.35 & \textbf{36.38}\\
		\hline
	\end{tabular}
	\label{inirefm}
\end{table}

\subsubsection{Enhancing the State-of-the-art WSSS Methods}

Using Region-CAM as the activation component in the WSSS pipeline by replacing the activation part of existing state-of-the-art algorithms, it not only enhances the performance, but also remains comparable to the most advanced WSSS algorithms \cite{chang2020weakly,wang2020self,sun2021ecs,chen2022self,xu2022multi,kweon2023weakly,chen2024spatial,zhao2024sfc} without requiring carefully designed network architectures or specialized training strategies to overcome its limitations . Table~\ref{tab:2} shows the mIoU of segmentation seeds generated by different WSSS methods, as well as by Region-CAM when integrated with the baseline and other frameworks. The baseline model is a pure classification network and is not specifically optimized for WSSS. The results show that Region-CAM combined with the baseline model outperforms the CNN-based WSSS methods in the literature. Furthermore, the performance of SIPE \cite{chen2022self} and MCTformer \cite{xu2022multi} was further improved by integrating them with the SIP mechanism of Region-CAM. SIPE and MCTformer were used because their weights were publicly available, ensuring fairness and reproducibility. This demonstrates the flexibility and the great potential of Region-CAM.

\begin{table}
	\centering
    \caption{The average value of mIoU over the threshold range [0,1]. The interval of threshold is 0.01.}
        \setlength{\tabcolsep}{1.5pt}
	\begin{tabular}{lc|ccccc}
    \hline
		\multicolumn{2}{c|}{Dataset} & CAM & Grad-CAM & Grad-CAM++ & LayerCAM & Region-CAM\\
        \hline
		\multirow{2}{*}{VOC} & Train & 26.78 & 25.64 & 26.75 & 26.59 & \textbf{38.39}\\
		& Val. & 26.40 & 25.19 & 25.94 & 26.35 & \textbf{37.67}\\
        \hline
	\end{tabular}
	\label{TheaveragevalueofmIoUinthethreshold}
\end{table}

Our activation method outperforms previous WSSS algorithms may be because the previous algorithm aims to improve the original CAM to alleviate its tendency to over-focus on the most discriminative regions of objects \cite{chang2020weakly,wang2020self,sun2021ecs,chen2022self,yang2025exploring}. Our method activates more complete object by fusing and propagating semantic information from shallow network layers, which alleviates the limitation of the CAM. In addition, the superpixel regions derived from shallow features are closely aligned with object boundaries, effectively suppressing the activation of background regions and obtaining more accurate segmentation seeds. More analysis details are given in the following Sec. \ref{AblationStudy}.

\subsubsection{Results Analysis and Ablation Study}
\label{AblationStudy}

\begin{table}[tb]
	\centering
    	\caption{Evaluation of the impact of semantic propagation based on different layers superpixel regions on the segmentation seeds results (The numbers of cluster centroids equals to 10).}
	\begin{tabular}{ccccc}
			     Block\_4 & Block\_3 & Block\_2 & Block\_1 & mIoU (\%)\\
			\hline          
			\checkmark & & & & 55.92\\
                &\checkmark& & & 56.74\\
                & & \checkmark & & 56.15\\
                & & & \checkmark & 54.10\\
                \hline
                    &\checkmark &\checkmark& & 57.90\\
                    &\checkmark & & \checkmark & 57.07\\
                    &\checkmark &\checkmark&\checkmark & 58.08\\
                \hline
                \checkmark & \checkmark & & & 58.66\\
                \checkmark & & \checkmark& & 57.99\\
                \checkmark & & &\checkmark & 57.07\\
                \checkmark &\checkmark&\checkmark& & 59.83\\
                \checkmark &\checkmark&\checkmark& \checkmark & \textbf{60.12}\\
                \hline
	\end{tabular}
	\label{tab:semanticpropagationsuperpixels}
\end{table}

Region-CAM achieves significant improvements for two key reasons: (1) The SIM, $S$, enriched by integrating semantic information from the shallow layers, captures more precise and detailed object information. Previous studies have shown that combining activation maps from different layers can lead to better performance \cite{jiang2021layercam,rebuffi2020there}. The difference is that Region-CAM obtains SIMs based on the non-negative gradients of each layer and merges these SIMs to obtain the $S$. (2) SIP process further extends semantics information from an initially small region to a broader region with superpixels generated by clustering network features. Since the features of spatial points in the same superpixel are highly similar. This ensures that the semantic information is reasonably expanded and not be propagated to non-target areas. In addition, the SIP mechanism from the deep layer to the shallow layer further ensures the correct propagation of semantic information. Although the deep layer features lack target details compared to the shallow layer features, they contain more global information. The clustering results based on the deep layer features can accurately separate the target from the background area.

Table \ref{inirefm}, shows the mIoU accuracy of the $S$ and the activation map $\bar{M}$. The accuracy of $S$ reaches 54.09\% on the VOC training set, which is 7.58\% higher than that for the CAM method. This proves that better activation maps can be achieved by leveraging shallow semantic information. Moreover, the SIP process further improves the result by 6.03\%. On the MS COCO validation set, $S$ reaches 32.35\%, which is 12.2\% higher than the original CAM. After the SIP process, the final activation map $\bar{M}$ achieves a segmentation accuracy of 36.38\%, which is 16.23\% higher than the original CAM. 

\begin{table}[tb]
	\centering
    	\caption{Compare the effects of different numbers of cluster centroids on the segmentation seed results. The layers are used from block\_1 to block\_4. as shown in Tab. \ref{tab:semanticpropagationsuperpixels}}
	\begin{tabular}{cccccc}
        \hline
			No. of centroids & 5 & 10 & 15  & 20 & 25\\
			\hline
			mIoU (\%) & 58.29& \textbf{60.12} &  59.42 & 58.85 & 58.20\\
        \hline
	\end{tabular}

	\label{tab:clustercentroid}
\end{table}

In addition, further ablation experiments were conducted to investigate the influence of superpixels extracted from different layers on the segmentation seed results, as presented in Table~\ref{tab:semanticpropagationsuperpixels}. The SIP process should ideally begin at Block\_4 of the network and progressively proceed toward shallower layers to achieve optimal performance. This can be attributed to the fact that the boundaries of clustering results of Block\_4 are more aligned with the object boundaries than those of Block\_5, and the clustering results of Block\_4 effectively separate the object from the background. This phenomenon can be clearly observed by comparing the results in column (h) of Figure \ref{semanticClustering} with the results in the other columns: (g), (i), (j) and (k). Importantly, this property is not specific to any single network architecture, but is consistently observed across multiple different models and datasets.

This misalignment of superpixels clustered from deepest features is likely attributed to the fact that each spatial point in this layer representations aggregates information from a broad receptive field. Deep features emphasize global contextual semantics rather than local structural details, leading to the loss of fine-grained information such as object boundaries \cite{chen2017deeplab}. In contrast, shallow-layer features exhibit opposite characteristics: they capture information from smaller local regions and preserve higher spatial resolution \cite{badrinarayanan2017segnet}. Therefore, in this study, the SIP mechanism is designed to initiate from the Block\_4 and progressively proceed toward the shallowest layer.

Moreover, the effects of different numbers of cluster centroids on the segmentation seed results are verified, as shown in Table~\ref{tab:clustercentroid}. The highest accuracy (60.12\%) is achieved with 10 centroids, indicating an optimal balance between object detail retention and spatial coherence. Performance declines when the number of centroids increases beyond 10, with 15, 20, and 25 centroids yielding progressively lower mIoU values (59.42\%, 58.85\%, and 58.20\%, respectively). This may be due to over segmentation leading to fragmented object structure, which is not beneficial for semantic propagation. Similarly, utilizing only 5 centroids results in 58.29\% mIoU. This result indicates that when segmentation is insufficient, object and background regions become difficult to distinguish, leading the semantic propagation process to extend semantics information into background areas and produce erroneous background activations. These results suggest that using 10 centroids provides the optimal balance between segmentation detail and noise suppression, whereas both excessive and insufficient clustering degrade segmentation accuracy.

\begin{figure*}[h]
    \centering
    \begin{minipage}{0.44\textwidth}
        \centering
        \includegraphics[width=\linewidth]{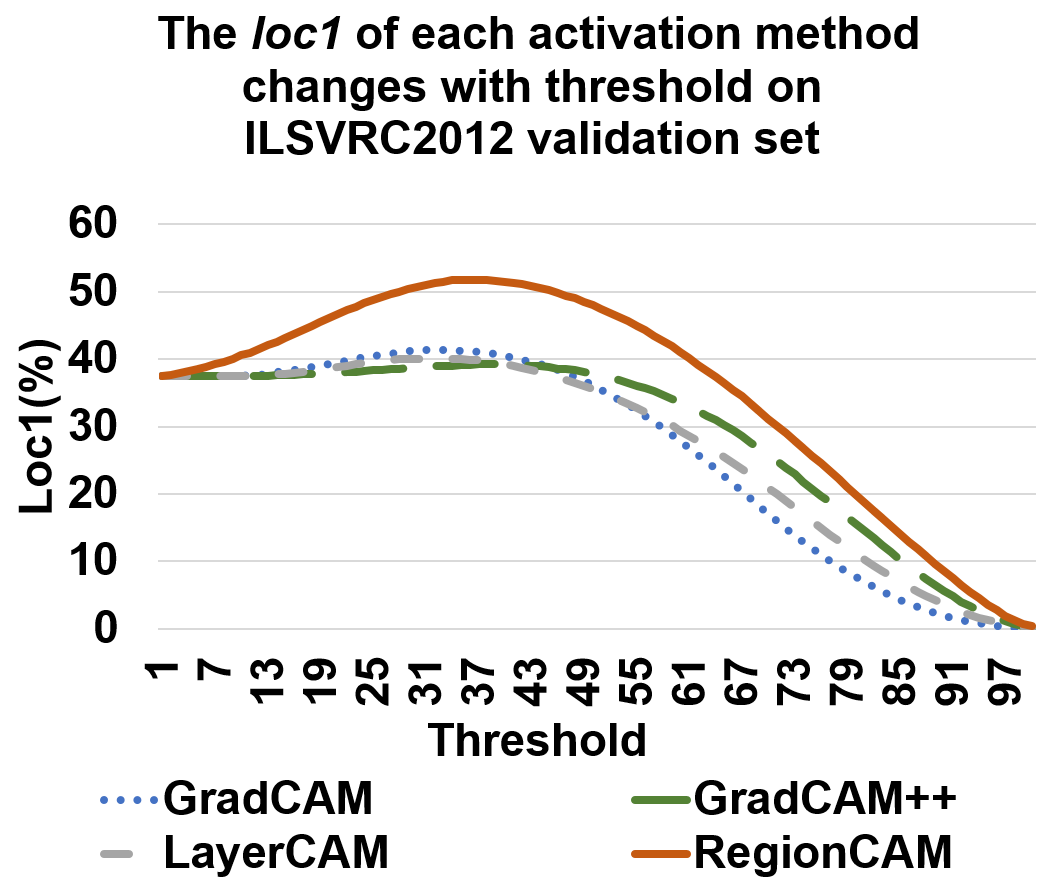}\\ (a) The loc1 changes with threshold \label{fig:short-b}
    \end{minipage}
    \begin{minipage}{0.44\textwidth}
        \centering
        \includegraphics[width=\linewidth]{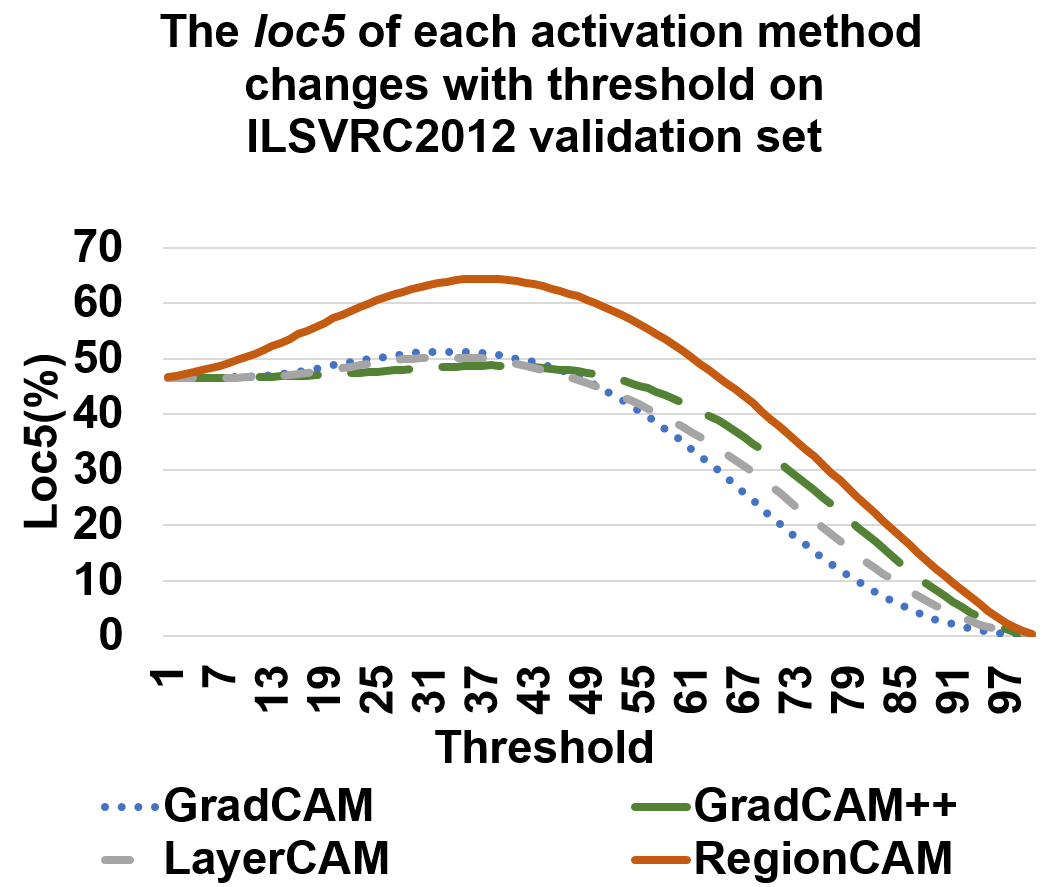}\\ (b) The loc5 changes with threshold
    \end{minipage}
    \caption{\textbf{The \textit{loc1} and \textit{loc5} of each activation method changes with threshold on the ILSVRC2012 validation set. The interval of threshold is 0.01.}}
    \label{loc1changewithThreshold}
\end{figure*}

\subsection{Weakly supervised object localization Experiment}
\label{Experiments1-2}
In this subsection, we further validate the superior performance of Region-CAM by evaluating it within an object localization experiment. Following the common practice \cite{wang2020score,selvaraju2017grad,ramaswamy2020ablation,jiang2021layercam}, we start this experiment by resizing the input images to 244 × 244 and normalizing them to the [0, 1] range using a mean vector of [0.485, 0.456, 0.406] and a standard deviation vector of [0.229, 0.224, 0.225].  For  fair comparison with previous methods, we use the pre-trained VGG16 network~\cite{simonyan2014very} from the PyTorch model zoo~\cite{pytorch} as our baseline model. The network consists of five stages, from which the semantic information maps (SIMs) and features are extracted from stages 4 to 1, respectively. These SIMs are processed following the procedure described in Subsection~\ref{Experiments1-1}. To generate predicted bounding boxes from Region-CAM maps, we set the threshold to 35\% of the maximum activation value to remove background regions and select the tightest bounding box around the largest connected segment as the predicted bounding box, following the common practice~\cite{zhou2016learning,jiang2021layercam}. 

We evaluate the performance by computing Top-1 localization accuracy (\textit{loc1}) and Top-5 localization accuracy (\textit{loc5}) on the ILSVRC2012 validation set~\cite{russakovsky2015imagenet}. The \textit{loc1} metric indicates that the top-1 predicted class is correct and that the Intersection over Union (IoU) between the predicted and ground-truth (GT) bounding boxes is at least 0.5, in which case the localization is considered correct.   The \textit{loc5} metric indicates that the GT class is among the top-5 predicted classes and that the corresponding bounding box also meets the IoU threshold, thereby being considered correctly localized \cite{zhou2016learning,jiang2021layercam}. 
It must be noted that the IoU used here comes from the field of object detection \cite{zou2023object}.
Although this IoU shares the same basic intersection-over-union principle with mIoU in the field of semantic segmentation, they are calculated differently.

\begin{figure}[]
    \centering
    \begin{minipage}{\textwidth} 
        \begin{minipage}{0.06\textwidth} 
            \raggedleft
            \rotatebox{90}{\scriptsize Input} 
        \end{minipage}%
        \hspace{0.01em}
        \begin{minipage}{0.88\textwidth} 
            \centering
            \includegraphics[width=\textwidth]{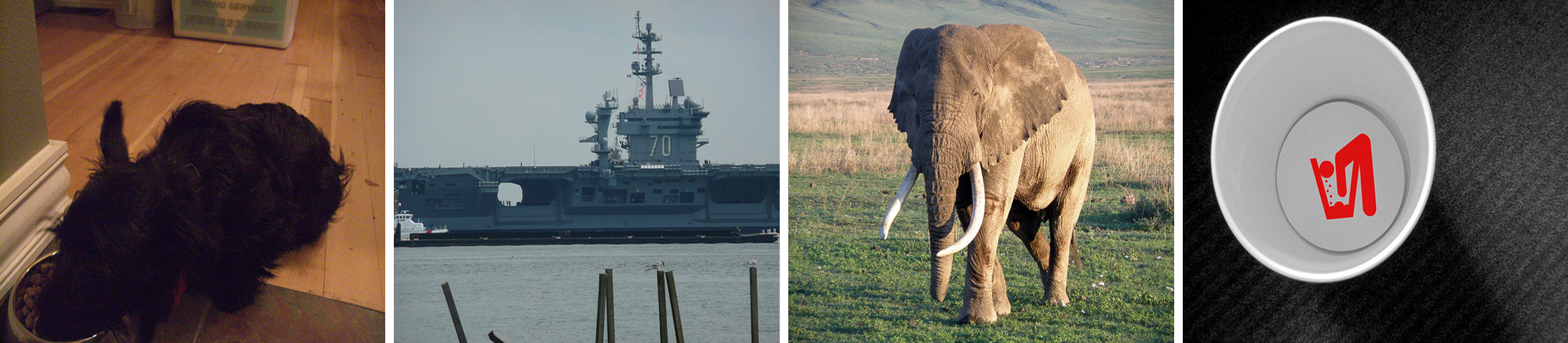} 
        \end{minipage}
    \end{minipage}

    \vspace{0.1em} 

    \begin{minipage}{\textwidth} 
        \begin{minipage}{0.06\textwidth} 
            \raggedleft
            \rotatebox{90}{\scriptsize Grad-CAM}
        \end{minipage}%
        \hspace{0.01em}
        \begin{minipage}{0.88\textwidth} 
            \centering
            \includegraphics[width=\textwidth]{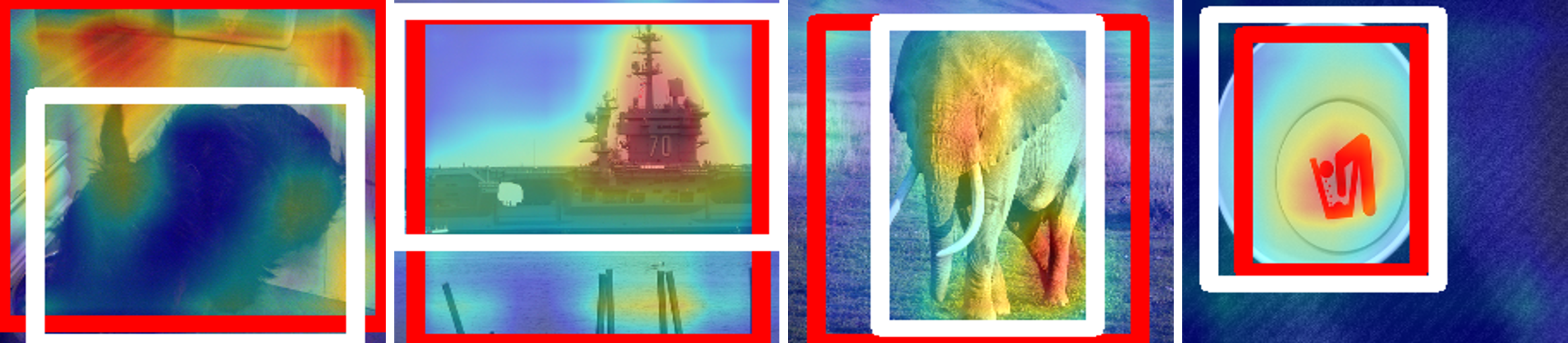}
        \end{minipage}
    \end{minipage}

    \vspace{0.1em} 

    \begin{minipage}{\textwidth} 
        \begin{minipage}{0.06\textwidth} 
            \raggedleft
            \rotatebox{90}{\scriptsize Grad-CAM++}
        \end{minipage}%
        \hspace{0.01em}
        \begin{minipage}{0.88\textwidth} 
            \centering
            \includegraphics[width=\textwidth]{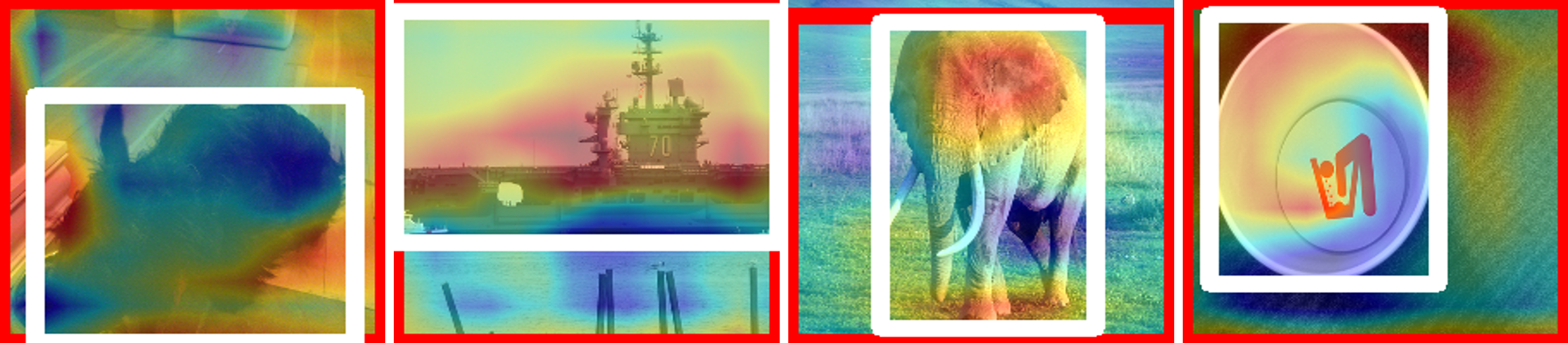}
        \end{minipage}
    \end{minipage}
    
    \vspace{0.1em} 

    \begin{minipage}{\textwidth} 
        \begin{minipage}{0.06\textwidth} 
            \raggedleft
            \rotatebox{90}{\scriptsize LayerCAM}
        \end{minipage}%
        \hspace{0.01em}
        \begin{minipage}{0.88\textwidth}
            \centering
            \includegraphics[width=\textwidth]{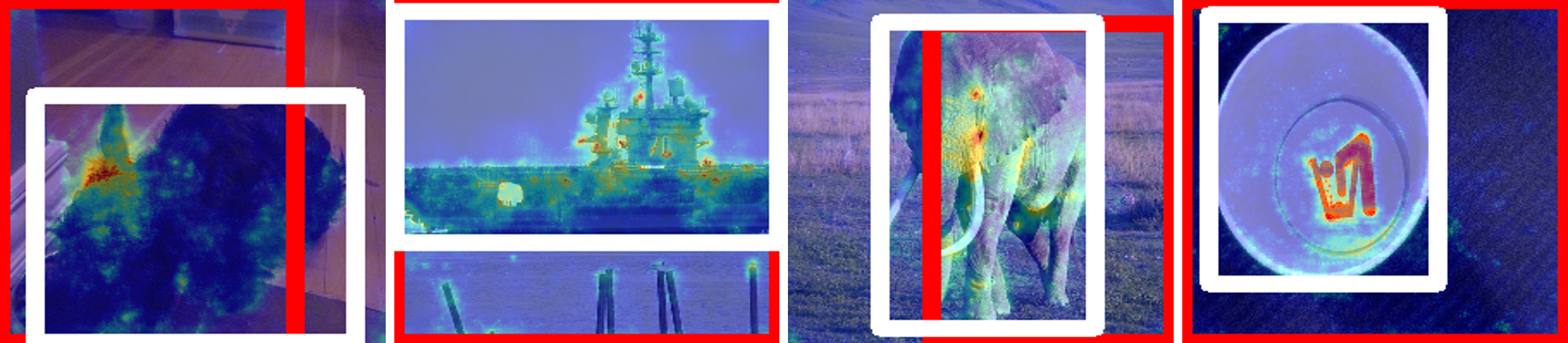}
        \end{minipage}
    \end{minipage}
    
    \vspace{0.1em} 
    
    \begin{minipage}{\textwidth} 
        \begin{minipage}{0.06\textwidth} 
            \raggedleft
            \rotatebox{90}{\scriptsize Region-CAM*}
        \end{minipage}%
        \hspace{0.01em}
        \begin{minipage}{0.88\textwidth} 
            \centering
            \includegraphics[width=\textwidth]{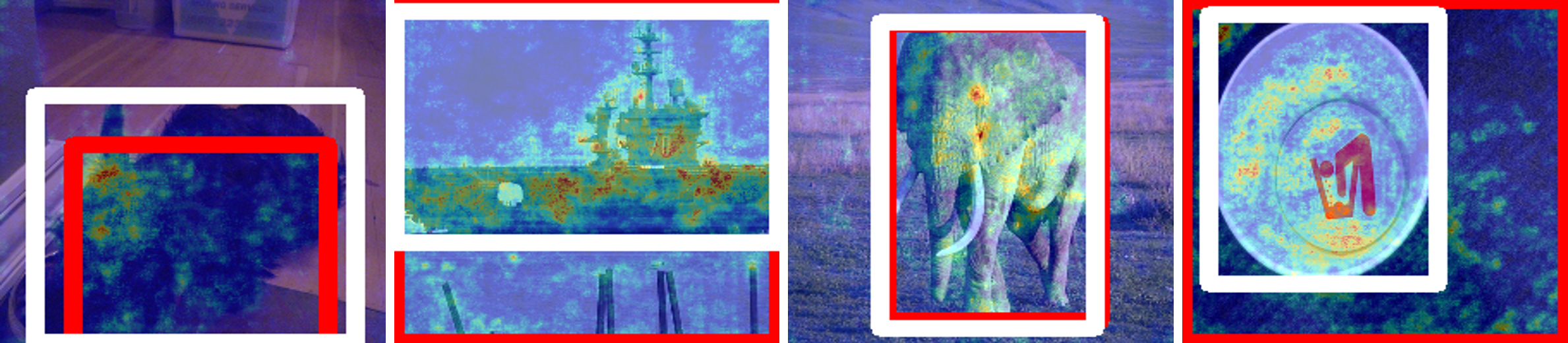}
        \end{minipage}
    \end{minipage}

    \vspace{0.1em} 
    
    \begin{minipage}{\textwidth} 
        \begin{minipage}{0.06\textwidth} 
            \raggedleft
            \rotatebox{90}{\scriptsize Region-CAM}
        \end{minipage}%
        \hspace{0.01em}
        \begin{minipage}{0.88\textwidth} 
            \centering
            \includegraphics[width=\textwidth]{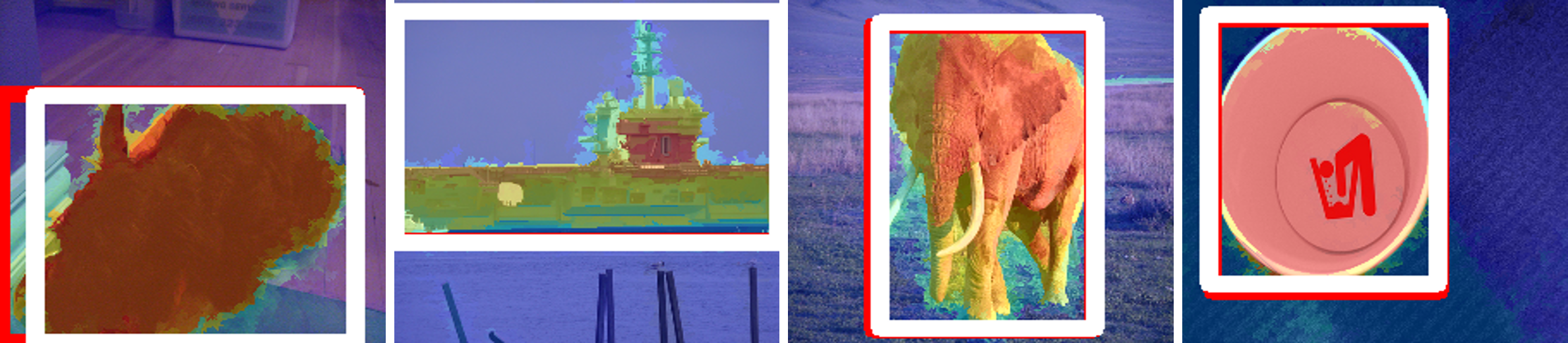}
        \end{minipage}
    \end{minipage}
    
    \caption{Qualitative comparison for localization results among different methods on ILSVRC validation set. Region-CAM* represents the $S$. Region-CAM means the class activation map $\bar{M}$. The white box is the GT box, and the red box refers to the predicted box.}
    \label{LOcresult}
\end{figure}

Compared with Grad-CAM, Grad-CAM++ and LayerCAm, our method performs best on \textit{loc1} and \textit{loc5}. As shown in Tab. \ref{tab:3}, the \textit{loc1} and \textit{loc5} reach 51.7\ and 64.3\%, which are 4.5\% and 5.6\% higher than the results of LayerCAM, respectively. In addition, the accuracy we obtained based on the $S$ is expressed in brackets. We can see that the result of $\bar{M}$ is much better than the $S$. This is mainly because the maximum pooling layer of the VGG16 network has an impact on the localization ability, which introduces visual noise \cite{zhou2016learning,wang2020score,ramaswamy2020ablation}. However, our SIP mechanism filters out these visual noises by averaging based on superpixel regions. This makes the activation map more accurate. As can be seen from the penultimate row of Fig. \ref{LOcresult}, there is a lot of noise in the activation map, but after SIP process, these noise points are eliminated in the last row. This further demonstrates the robustness of the SIP mechanism of our algorithm.

\begin{table}[tb]
  \centering
    \caption{Comparison of the localization performance of our method with other methods based on \textit{loc1} and \textit{loc5}. * Indicates result is from the original paper. The results in the brackets are $S$ \cite{jiang2021layercam}.}
    \setlength{\tabcolsep}{2pt}
  \begin{tabular}{lcccc}
    \hline
    Method & Grad-CAM* & Grad-CAM++* & LayerCAM* & Region-CAM \\
    \hline
    \textit{loc1} (\%) & 43.6 & 45.4 & 47.2 & \textbf{51.7}(43.1)\\
    \textit{loc5} (\%) & 54.0 & 56.4 & 58.7 & \textbf{64.3}(53.7)\\
    \hline
  \end{tabular}
  \label{tab:3}
\end{table}

Similar to the generation of semantic seeds, the localization results are influenced by the threshold. Therefore, the variations in \textit{loc1} and \textit{loc5} with respect to different threshold values are shown in Figure \ref{loc1changewithThreshold}. Region-CAM consistently outperforms the other three methods in localization, regardless of the threshold applied. The average \textit{loc1} and \textit{loc5} of each algorithm over the threshold [0,1] range is shown in Table.~\ref{averageloc1andloc5}.

\begin{table}[tb]
	\centering
    \caption{The average \textit{loc1} and \textit{loc5} over the threshold range [0,1] on the ILSVRC2012 validation set. The higher the value, the better.}
	\label{averageloc1andloc5}
    \setlength{\tabcolsep}{2pt}
	\begin{tabular}{lcccc}
			\hline
			Method & Grad-CAM & Grad-CAM++ & LayerCAM & Region-CAM \\
			\hline
			Ave.\textit{loc1} (\%) & 26.62 & 28.90 & 27.31 & \textbf{35.50} \\
			Ave. \textit{loc5} (\%) & 33.26 & 36.08 & 34.48 & \textbf{44.32} \\
			\hline
	\end{tabular}
\end{table}

\begin{table*}[t]
	\centering
    \caption{The image occlusion results of each algorithm. Original represents the Top-1 and Top-5 accuracy (\%) of VGG16 without occlusion. Image occlusion results of each algorithm. After image occlusion, the lower the accuracy and confidence, the stronger the algorithm's explanatory power. Therfore, the lower the accuracy and confidence, the better.}
	\begin{tabular}{lcccc}
			Method & Top-1 accuracy (\%)  & Top-5 accuracy (\%) & Top-1 confidence (\%) & Top-5 confidence (\%) \\
			\hline
			Original & 70.02 &  89.41 &  83.81 & 69.04 \\
			Grad-CAM  &  62.53 &  84.23 & 79.51 & 62.92\\
			Grad-CAM++ & 65.31 & 85.63 & 81.11 & 65.54\\
			LayerCAM & 68.45 &  88.05 & 82.88 & 67.92 \\
			Ours Region-CAM & \textbf{57.16} & \textbf{79.33} & \textbf{78.68} & \textbf{60.74} \\
            \hline
	\end{tabular}
	\label{tab:4}
\end{table*}

\subsection{Image Occlusion Experiment}
\label{Experiments1-3}

In this subsection we present another experiment to demonstrate the usefulness of Region-CAM. In this image occlusion experiment, we aim to evaluate the activation capabilities of CAM algorithms by systematically occluding portions of the input image and observing the resulting changes in model performance \cite{zeiler2014visualizing,jiang2021layercam}. 

The validation experiments presented in previous subsections (segmentation seed generation and weakly supervised object localization) evaluate the performance of activation algorithms from an application-oriented perspective, primarily measuring how activation maps contribute to improving the accuracy and reliability of downstream tasks, such as, WSSS or WSOL. In other words, they reflect the practical benefits of integrating activation algorithms into end-to-end weakly supervised learning pipelines. However, image-level WSSS and WSOL models are inherently classification networks, and activation algorithms serve as mechanisms for processing and interpreting the feature representations of these networks. 

On the other hand, the image occlusion experiments presented in this subsection, which evaluate the capabilities of the class activation maps based on the network classification results, provide a direct and closer-to-the-model-nature approach. This experiment assesses the activation algorithm’s intrinsic ability to capture and represent the object regions most critical to the network’s predictions. Moreover, occlusion-based evaluation offers additional evidence of the activation algorithm’s reliability and generalization capability. Accordingly, we conducted image occlusion experiments to verify Region-CAM’s ability to capture and represent object regions, thereby further validating its reliability and generalization performance.

Specifically, portions of the image corresponding to regions of high activation are gradually occluded, and the Top-1 and Top-5 classification accuracy and prediction confidence are recalculated. If occluding an activated region causes a significant decrease in classification accuracy or confidence, it indicates that the activation map has correctly identified regions that are critical to the model's decision. Conversely, slight changes after occlusion indicate that the highlighted region may not have a significant impact on the prediction. At the same occlusion threshold, a greater decrease in this metric indicates better activation performance. 

\begin{figure}[h]
    \centering
    \begin{minipage}{0.14\textwidth}
        \centering
        \includegraphics[width=\linewidth]{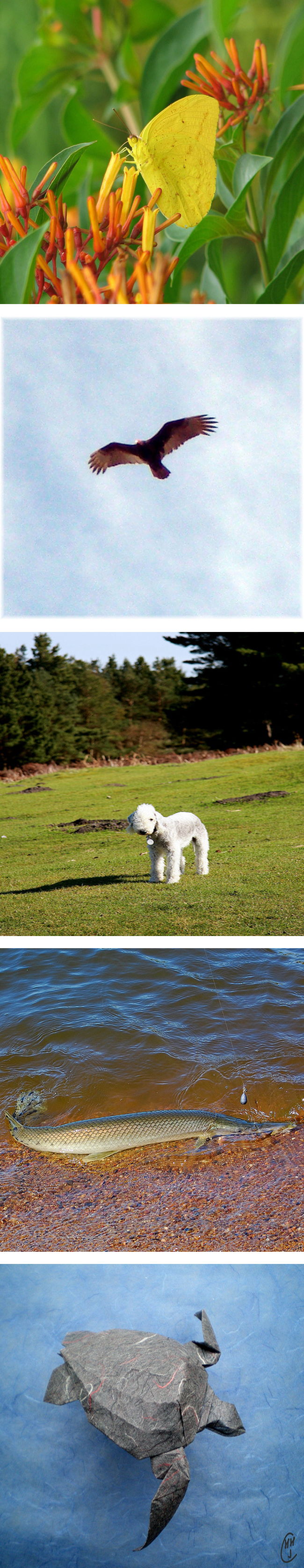}\\ Original image
    \end{minipage}
    \begin{minipage}{0.14\textwidth}
        \centering
        \includegraphics[width=\linewidth]{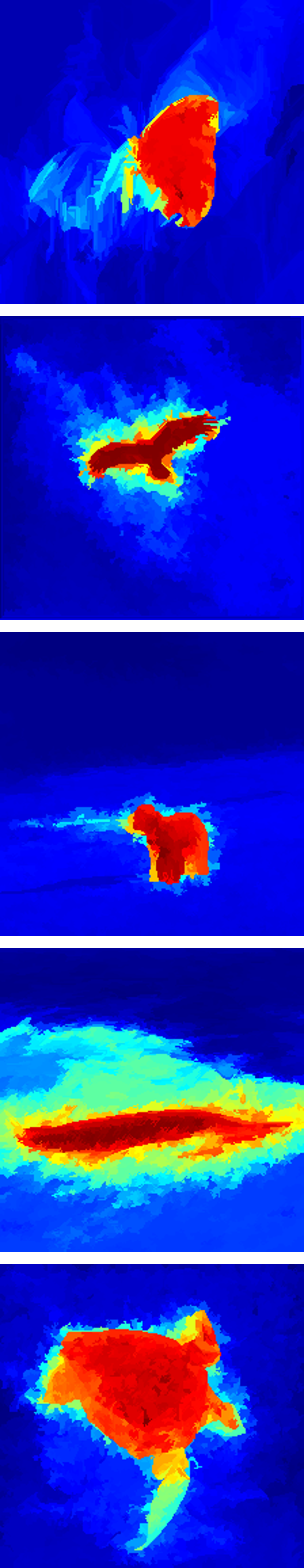}\\  Activation map
    \end{minipage}
    \begin{minipage}{0.14\textwidth}
        \centering
        \includegraphics[width=\linewidth]{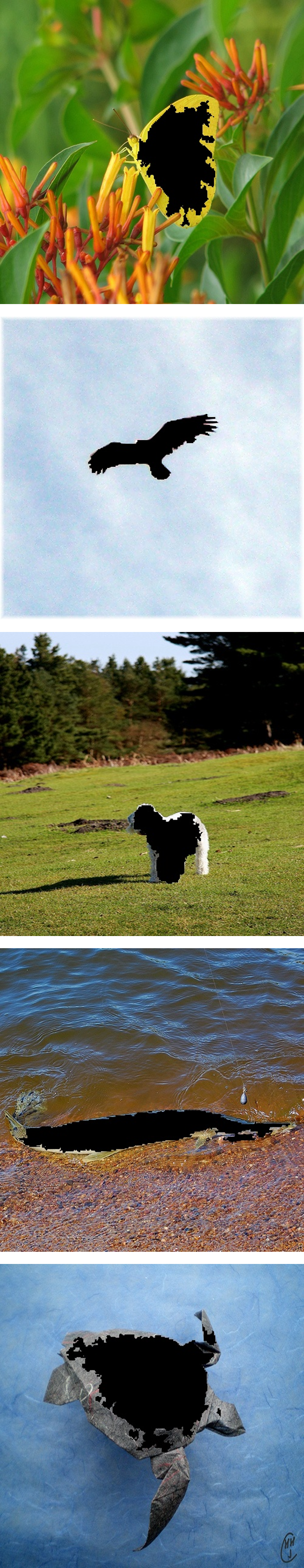}\\ Occluded object
    \end{minipage}
    \caption{\textbf{An original image of a target class, followed by its activation map obtained from Region-CAM, and the corresponding occlusion map.}}
    \label{Occlusion}
\end{figure}

Image occlusion experiments were conducted using the VGG16 model on the ILSVRC2012 validation set. Image regions with confidence exceeding 85\% of the maximum were occluded, after which the occluded image was fed into the network to recompute the Top-1 and Top-5 accuracy and confidence. Figure~\ref{Occlusion} shows the original image, the corresponding activation map, and the occlusion image generated by Region-CAM.

The results of the image occlusion experiment is shown in Table~\ref{tab:4}. Compared with other methods, our Region-CAM has the largest decrease in the Top-1 and Top-5 accuracy and confidence. When masking the image, it can be observed that removing the confidence regions identified by Region-CAM leads to a more significant reduction in prediction scores compared to those of Grad-CAM, Grad-CAM++, and LayerCAM. This demonstrates that the class activation maps generated by Region-CAM can more accurately identify the spatial regions of an object that are most important for the target category. These results further verify the reliability of the confidence regions localized by Region-CAM.

\section{Conclusions}
\label{sec:conclusions}
In this work, we have proposes a novel calss activation mapping method, called Region-CAM, that benefits weakly supervised learning tasks, such as, WSSS and WSOL. Different from the previous weighted feature method, we considered the visual information provided by the gradients and features separately to maximize the effectiveness. More comprehensive information of the target can be captured through SIMs calculated by non-negative gradients of different layers. Moreover, The SIP mechanism, designed based on the superpixel regions clustered by different layers features, further removes noise in the SIM leading to accurate pixel-level class activation maps. 

Region-CAM outperforms other widely used activation methods in generating segmentation seeds, in object localization and in activating object-related regions. In terms of pixel-level accuracy, it has achieved 60.12\% and 58.43\% mean intersection over union (mIoU) using the baseline model on the PASCAL VOC training and validation datasets, respectively.  This is an improvement of 13.61\% and 13.13\%, respectively over the original CAM (46.51\% and 45.30\%). On the MS COCO validation set, our method achieved 36.38\%, corresponding to a 16.23\% improvement over the CAM (20.15\%). Moreover, when the class activation components of other WSSS algorithms are replaced with our Region-CAM, the accuracy of the seeds generated by those algorithms is further improved. By replacing the activation components of SIPE with our Region-CAM, the results improved from 58.6\% to 62.8\%. In addition,  the CAM method in MCTformer is replaced with Region-CAM, the results improved from 61.7\% to 64.2\%, which is comparable to the state-of-the-art algorithms. For localization, Region-CAM achieved 51.7\% in \textit{Loc1} and 64.3\% in \textit{Loc5} on the ILSVRC2012 validation set. Compared to LayerCAM, our algorithm improves performance by 4.5\% in \textit{loc1} and 5.6\% in \textit{loc5}. 

Overall, Region-CAM generates more accurate and complete class activation maps. While maintaining precise activation region boundaries, it also highlights a larger portion of the target regions. This indicates that the proposed method effectively leverages information from conventional classification networks to improve the quality of weakly supervised segmentation seeds and object localization results, thereby laying a solid foundation for downstream weakly supervised learning tasks.

\bibliographystyle{unsrt}
\bibliography{mybib}  

\begin{thebibliography}{10}

\bibitem{zhou2016learning}
Bolei Zhou, Aditya Khosla, Agata Lapedriza, Aude Oliva, and Antonio Torralba.
\newblock Learning deep features for discriminative localization.
\newblock In {\em Proceedings of the IEEE conference on computer vision and pattern recognition}, pages 2921--2929, 2016.

\bibitem{selvaraju2017grad}
Ramprasaath~R Selvaraju, Michael Cogswell, Abhishek Das, Ramakrishna Vedantam, Devi Parikh, and Dhruv Batra.
\newblock Grad-\uppercase{cam}: Visual explanations from deep networks via gradient-based localization.
\newblock In {\em Proceedings of the IEEE international conference on computer vision}, pages 618--626, 2017.

\bibitem{wang2020score}
Haofan Wang, Zifan Wang, Mengnan Du, Fan Yang, Zijian Zhang, Sirui Ding, Piotr Mardziel, and Xia Hu.
\newblock Score-\uppercase{CAM}: Score-weighted visual explanations for convolutional neural networks.
\newblock In {\em Proceedings of the IEEE/CVF conference on computer vision and pattern recognition workshops}, pages 24--25, 2020.

\bibitem{zhang2020survey}
Man Zhang, Yong Zhou, Jiaqi Zhao, Yiyun Man, Bing Liu, and Rui Yao.
\newblock A survey of semi-and weakly supervised semantic segmentation of images.
\newblock {\em Artificial Intelligence Review}, 53:4259--4288, 2020.

\bibitem{li2018weakly}
Qizhu Li, Anurag Arnab, and Philip~HS Torr.
\newblock Weakly-and semi-supervised panoptic segmentation.
\newblock In {\em Proceedings of the European conference on computer vision (ECCV)}, pages 102--118, 2018.

\bibitem{li2023weakly}
Jinlong Li, Zequn Jie, Xu~Wang, Yu~Zhou, Lin Ma, and Jianmin Jiang.
\newblock Weakly supervised semantic segmentation via self-supervised destruction learning.
\newblock {\em Neurocomputing}, 561:126821, 2023.

\bibitem{ru2022weakly}
Lixiang Ru, Bo~Du, Yibing Zhan, and Chen Wu.
\newblock Weakly-supervised semantic segmentation with visual words learning and hybrid pooling.
\newblock {\em International Journal of Computer Vision}, 130(4):1127--1144, 2022.

\bibitem{alom2018recurrent}
Md~Zahangir Alom, Mahmudul Hasan, Chris Yakopcic, Tarek~M Taha, and Vijayan~K Asari.
\newblock Recurrent residual convolutional neural network based on u-net (r2u-net) for medical image segmentation.
\newblock {\em arXiv preprint arXiv:1802.06955}, 2018.

\bibitem{chen2017deeplab}
Liang-Chieh Chen, George Papandreou, Iasonas Kokkinos, Kevin Murphy, and Alan~l Yuille.
\newblock Deep\uppercase{l}ab: Semantic image segmentation with deep convolutional nets, atrous convolution, and fully connected crfs.
\newblock {\em IEEE transactions on pattern analysis and machine intelligence}, 40(4):834--848, 2017.

\bibitem{chen2018encoder}
Liang-Chieh Chen, Yukun Zhu, George Papandreou, Florian Schroff, and Hartwig Adam.
\newblock Encoder-decoder with atrous separable convolution for semantic image segmentation.
\newblock In {\em Proceedings of the European conference on computer vision (ECCV)}, pages 801--818, 2018.

\bibitem{cheng2022masked}
Bowen Cheng, Ishan Misra, Alexander~G Schwing, Alexander Kirillov, and Rohit Girdhar.
\newblock Masked-attention mask transformer for universal image segmentation.
\newblock In {\em Proceedings of the IEEE/CVF conference on computer vision and pattern recognition}, pages 1290--1299, 2022.

\bibitem{xu2022cream}
Jilan Xu, Junlin Hou, Yuejie Zhang, Rui Feng, Rui-Wei Zhao, Tao Zhang, Xuequan Lu, and Shang Gao.
\newblock Cream: Weakly supervised object localization via class re-activation mapping.
\newblock In {\em Proceedings of the IEEE/CVF conference on computer vision and pattern recognition}, pages 9437--9446, 2022.

\bibitem{choe2020evaluating}
Junsuk Choe, Seong~Joon Oh, Seungho Lee, Sanghyuk Chun, Zeynep Akata, and Hyunjung Shim.
\newblock Evaluating weakly supervised object localization methods right.
\newblock In {\em Proceedings of the IEEE/CVF conference on computer vision and pattern recognition}, pages 3133--3142, 2020.

\bibitem{zhang2021weakly}
Dingwen Zhang, Junwei Han, Gong Cheng, and Ming-Hsuan Yang.
\newblock Weakly supervised object localization and detection: A survey.
\newblock {\em IEEE transactions on pattern analysis and machine intelligence}, 44(9):5866--5885, 2021.

\bibitem{kim2021normalization}
Jeesoo Kim, Junsuk Choe, Sangdoo Yun, and Nojun Kwak.
\newblock Normalization matters in weakly supervised object localization.
\newblock In {\em Proceedings of the IEEE/CVF international conference on computer vision}, pages 3427--3436, 2021.

\bibitem{wang2020weakly}
Xiang Wang, Sifei Liu, Huimin Ma, and Ming-Hsuan Yang.
\newblock Weakly-supervised semantic segmentation by iterative affinity learning.
\newblock {\em International Journal of Computer Vision}, 128:1736--1749, 2020.

\bibitem{cai2023ssdb}
Qingdong Cai and Charith Abhayaratne.
\newblock S\uppercase{SDB}-\uppercase{N}et: A single-step dual branch network for weakly supervised semantic segmentation of food images.
\newblock In {\em 2023 IEEE 25th International Workshop on Multimedia Signal Processing (MMSP)}, pages 1--6. IEEE, 2023.

\bibitem{wang2017weakly}
Yu~Wang, Fengqing Zhu, Carol~J Boushey, and Edward~J Delp.
\newblock Weakly supervised food image segmentation using class activation maps.
\newblock In {\em 2017 IEEE International Conference on Image Processing (ICIP)}, pages 1277--1281. IEEE, 2017.

\bibitem{shimoda2020weakly}
Wataru Shimoda and Keiji Yanai.
\newblock Weakly-supervised plate and food region segmentation.
\newblock In {\em 2020 IEEE International Conference on Multimedia and Expo (ICME)}, pages 1--6. IEEE, 2020.

\bibitem{yang2025exploring}
Zhiwei Yang, Yucong Meng, Kexue Fu, Feilong Tang, Shuo Wang, and Zhijian Song.
\newblock Exploring clip's dense knowledge for weakly supervised semantic segmentation.
\newblock In {\em Proceedings of the Computer Vision and Pattern Recognition Conference}, pages 20223--20232, 2025.

\bibitem{chen2022class}
Zhaozheng Chen, Tan Wang, Xiongwei Wu, Xian-Sheng Hua, Hanwang Zhang, and Qianru Sun.
\newblock Class re-activation maps for weakly-supervised semantic segmentation.
\newblock In {\em Proceedings of the IEEE/CVF conference on computer vision and pattern recognition}, pages 969--978, 2022.

\bibitem{chou2017framecnn}
Szu-Yu Chou, Jyh-Shing~Roger Jang, and Yi-Hsuan Yang.
\newblock Framecnn: A weakly-supervised learning framework for frame-wise acoustic event detection and classification.
\newblock {\em Recall}, 14:55--64, 2017.

\bibitem{wu2025prompt}
Wangyu Wu, Xianglin Qiu, Siqi Song, Zhenhong Chen, Xiaowei Huang, Fei Ma, and Jimin Xiao.
\newblock Prompt categories cluster for weakly supervised semantic segmentation.
\newblock In {\em Proceedings of the Computer Vision and Pattern Recognition Conference}, pages 3198--3207, 2025.

\bibitem{xu2019improved}
Lian Xu, Mohammed Bennamoun, Farid Boussa{\"\i}d, Senjian An, and Ferdous Sohel.
\newblock An improved approach to weakly supervised semantic segmentation.
\newblock In {\em ICASSP 2019-2019 IEEE International Conference on Acoustics, Speech and Signal Processing (ICASSP)}, pages 1897--1901. IEEE, 2019.

\bibitem{zhang2020self}
Xiao Zhang and Michael Maire.
\newblock Self-supervised visual representation learning from hierarchical grouping.
\newblock {\em Advances in Neural Information Processing Systems}, 33:16579--16590, 2020.

\bibitem{chattopadhay2018grad}
Aditya Chattopadhay, Anirban Sarkar, Prantik Howlader, and Vineeth~N Balasubramanian.
\newblock Grad-\uppercase{cam}++: Generalized gradient-based visual explanations for deep convolutional networks.
\newblock In {\em 2018 IEEE winter conference on applications of computer vision (WACV)}, pages 839--847. IEEE, 2018.

\bibitem{ahn2019weakly}
Jiwoon Ahn, Sunghyun Cho, and Suha Kwak.
\newblock Weakly supervised learning of instance segmentation with inter-pixel relations.
\newblock In {\em Proceedings of the IEEE/CVF Conference on Computer Vision and Pattern Recognition}, pages 2209--2218, 2019.

\bibitem{ahn2018learning}
Jiwoon Ahn and Suha Kwak.
\newblock Learning pixel-level semantic affinity with image-level supervision for weakly supervised semantic segmentation.
\newblock In {\em Proceedings of the IEEE Conference on Computer Vision and Pattern Recognition}, pages 4981--4990, 2018.

\bibitem{chen2020weakly}
Liyi Chen, Weiwei Wu, Chenchen Fu, Xiao Han, and Yuntao Zhang.
\newblock Weakly supervised semantic segmentation with boundary exploration.
\newblock In {\em European Conference on Computer Vision}, pages 347--362. Springer, 2020.

\bibitem{jiang2021layercam}
Peng-Tao Jiang, Chang-Bin Zhang, Qibin Hou, Ming-Ming Cheng, and Yunchao Wei.
\newblock Layer\uppercase{cam}: Exploring hierarchical class activation maps for localization.
\newblock {\em IEEE Transactions on Image Processing}, 30:5875--5888, 2021.

\bibitem{shi2021zoom}
Xiangwei Shi, Seyran Khademi, Yunqiang Li, and Jan van Gemert.
\newblock Zoom-\uppercase{cam}: Generating fine-grained pixel annotations from image labels.
\newblock In {\em 2020 25th International Conference on Pattern Recognition (ICPR)}, pages 10289--10296. IEEE, 2021.

\bibitem{rebuffi2020there}
Sylvestre-Alvise Rebuffi, Ruth Fong, Xu~Ji, and Andrea Vedaldi.
\newblock There and back again: Revisiting backpropagation saliency methods.
\newblock In {\em Proceedings of the IEEE/CVF Conference on Computer Vision and Pattern Recognition}, pages 8839--8848, 2020.

\bibitem{yasuki2024cam}
Shunsuke Yasuki and Masato Taki.
\newblock \uppercase{CAM} back again: Large kernel cnns from a weakly supervised object localization perspective.
\newblock In {\em Proceedings of the IEEE/CVF Conference on Computer Vision and Pattern Recognition}, pages 341--351, 2024.

\bibitem{omeiza2019smooth}
Daniel Omeiza, Skyler Speakman, Celia Cintas, and Komminist Weldermariam.
\newblock Smooth \uppercase{g}rad-\uppercase{cam}++: An enhanced inference level visualization technique for deep convolutional neural network models.
\newblock {\em arXiv preprint arXiv:1908.01224}, 2019.

\bibitem{bany2021eigen}
Mohammed Bany~Muhammad and Mohammed Yeasin.
\newblock Eigen-\uppercase{CAM}: Visual explanations for deep convolutional neural networks.
\newblock {\em SN Computer Science}, 2(1):47, 2021.

\bibitem{ramaswamy2020ablation}
Harish~Guruprasad Ramaswamy et~al.
\newblock Ablation-cam: Visual explanations for deep convolutional network via gradient-free localization.
\newblock In {\em Proceedings of the IEEE/CVF Winter Conference on Applications of Computer Vision}, pages 983--991, 2020.

\bibitem{zhang2021group}
Qinglong Zhang, Lu~Rao, and Yubin Yang.
\newblock Group-\uppercase{CAM}: Group score-weighted visual explanations for deep convolutional networks.
\newblock {\em arXiv preprint arXiv:2103.13859}, 2021.

\bibitem{chen2022score}
Yifan Chen and Guoqiang Zhong.
\newblock Score-\uppercase{CAM}++: Class discriminative localization with feature map selection.
\newblock In {\em Journal of Physics: Conference Series}, volume 2278, page 012018. IOP Publishing, 2022.

\bibitem{yang2024decomcam}
Yuguang Yang, Runtang Guo, Sheng Wu, Yimi Wang, Linlin Yang, Bo~Fan, Jilong Zhong, Juan Zhang, and Baochang Zhang.
\newblock Decom\uppercase{CAM}: Advancing beyond saliency maps through decomposition and integration.
\newblock {\em Neurocomputing}, page 127826, 2024.

\bibitem{wu2021embedded}
Tong Wu, Junshi Huang, Guangyu Gao, Xiaoming Wei, Xiaolin Wei, Xuan Luo, and Chi~Harold Liu.
\newblock Embedded discriminative attention mechanism for weakly supervised semantic segmentation.
\newblock In {\em Proceedings of the IEEE/CVF conference on computer vision and pattern recognition}, pages 16765--16774, 2021.

\bibitem{meng2021foreground}
Meng Meng, Tianzhu Zhang, Qi~Tian, Yongdong Zhang, and Feng Wu.
\newblock Foreground activation maps for weakly supervised object localization.
\newblock In {\em Proceedings of the IEEE/CVF international conference on computer vision}, pages 3385--3395, 2021.

\bibitem{chen2022self}
Qi~Chen, Lingxiao Yang, Jian-Huang Lai, and Xiaohua Xie.
\newblock Self-supervised image-specific prototype exploration for weakly supervised semantic segmentation.
\newblock In {\em Proceedings of the IEEE/CVF conference on computer vision and pattern recognition}, pages 4288--4298, 2022.

\bibitem{chang2020weakly}
Yu-Ting Chang, Qiaosong Wang, Wei-Chih Hung, Robinson Piramuthu, Yi-Hsuan Tsai, and Ming-Hsuan Yang.
\newblock Weakly-supervised semantic segmentation via sub-category exploration.
\newblock In {\em Proceedings of the IEEE/CVF Conference on Computer Vision and Pattern Recognition}, pages 8991--9000, 2020.

\bibitem{wang2020self}
Yude Wang, Jie Zhang, Meina Kan, Shiguang Shan, and Xilin Chen.
\newblock Self-supervised equivariant attention mechanism for weakly supervised semantic segmentation.
\newblock In {\em Proceedings of the IEEE/CVF Conference on Computer Vision and Pattern Recognition}, pages 12275--12284, 2020.

\bibitem{lee2021anti}
Jungbeom Lee, Eunji Kim, and Sungroh Yoon.
\newblock Anti-adversarially manipulated attributions for weakly and semi-supervised semantic segmentation.
\newblock In {\em Proceedings of the IEEE/CVF conference on computer vision and pattern recognition}, pages 4071--4080, 2021.

\bibitem{kweon2021unlocking}
Hyeokjun Kweon, Sung-Hoon Yoon, Hyeonseong Kim, Daehee Park, and Kuk-Jin Yoon.
\newblock Unlocking the potential of ordinary classifier: Class-specific adversarial erasing framework for weakly supervised semantic segmentation.
\newblock In {\em Proceedings of the IEEE/CVF international conference on computer vision}, pages 6994--7003, 2021.

\bibitem{wu2024masked}
Fangwen Wu, Jingxuan He, Yufei Yin, Yanbin Hao, Gang Huang, and Lechao Cheng.
\newblock Masked collaborative contrast for weakly supervised semantic segmentation.
\newblock In {\em Proceedings of the IEEE/CVF winter conference on applications of computer vision}, pages 862--871, 2024.

\bibitem{shao2024knowledge}
Feifei Shao, Yawei Luo, Fei Gao, Yi~Yang, and Jun Xiao.
\newblock Knowledge-guided causal intervention for weakly-supervised object localization.
\newblock {\em IEEE Transactions on Knowledge and Data Engineering}, 2024.

\bibitem{wang2023treating}
Changwei Wang, Rongtao Xu, Shibiao Xu, Weiliang Meng, and Xiaopeng Zhang.
\newblock Treating pseudo-labels generation as image matting for weakly supervised semantic segmentation.
\newblock In {\em Proceedings of the IEEE/CVF International Conference on Computer Vision}, pages 755--765, 2023.

\bibitem{zhang2021complementary}
Fei Zhang, Chaochen Gu, Chenyue Zhang, and Yuchao Dai.
\newblock Complementary patch for weakly supervised semantic segmentation.
\newblock In {\em Proceedings of the IEEE/CVF International Conference on Computer Vision}, pages 7242--7251, 2021.

\bibitem{yang2024foundation}
Xiaobo Yang and Xiaojin Gong.
\newblock Foundation model assisted weakly supervised semantic segmentation.
\newblock In {\em Proceedings of the IEEE/CVF winter conference on applications of computer vision}, pages 523--532, 2024.

\bibitem{xu2025weakly}
Xiangfeng Xu, Pinyi Zhang, Wenxuan Huang, Yunhang Shen, Haosheng Chen, Jingzhong Lin, Wei Li, Gaoqi He, Jiao Xie, and Shaohui Lin.
\newblock Weakly supervised semantic segmentation via progressive confidence region expansion.
\newblock In {\em Proceedings of the Computer Vision and Pattern Recognition Conference}, pages 9829--9838, 2025.

\bibitem{zhu2025weakclip}
Lianghui Zhu, Xinggang Wang, Jiapei Feng, Tianheng Cheng, Yingyue Li, Bo~Jiang, Dingwen Zhang, and Junwei Han.
\newblock Weak\uppercase{clip}: Adapting clip for weakly-supervised semantic segmentation.
\newblock {\em International Journal of Computer Vision}, 133(3):1085--1105, 2025.

\bibitem{chen2024adaptive}
Zhiwei Chen, Siwei Wang, Liujuan Cao, Yunhang Shen, and Rongrong Ji.
\newblock Adaptive zone learning for weakly supervised object localization.
\newblock {\em IEEE Transactions on Neural Networks and Learning Systems}, 2024.

\bibitem{kumar2017hide}
Krishna Kumar~Singh and Yong Jae~Lee.
\newblock Hide-and-seek: Forcing a network to be meticulous for weakly-supervised object and action localization.
\newblock In {\em Proceedings of the IEEE international conference on computer vision}, pages 3524--3533, 2017.

\bibitem{xue2019danet}
Haolan Xue, Chang Liu, Fang Wan, Jianbin Jiao, Xiangyang Ji, and Qixiang Ye.
\newblock Da\uppercase{n}et: Divergent activation for weakly supervised object localization.
\newblock In {\em Proceedings of the IEEE/CVF International Conference on Computer Vision}, pages 6589--6598, 2019.

\bibitem{xie2021online}
Jinheng Xie, Cheng Luo, Xiangping Zhu, Ziqi Jin, Weizeng Lu, and Linlin Shen.
\newblock Online refinement of low-level feature based activation map for weakly supervised object localization.
\newblock In {\em Proceedings of the IEEE/CVF international conference on computer vision}, pages 132--141, 2021.

\bibitem{everingham2010pascal}
Mark Everingham, Luc Van~Gool, Christopher~KI Williams, John Winn, and Andrew Zisserman.
\newblock The \uppercase{PASCAL} \uppercase{V}isual \uppercase{O}bject \uppercase{C}lasses \uppercase{(VOC)} challenge.
\newblock {\em International journal of computer vision}, 88(2):303--338, 2010.

\bibitem{wu2019wider}
Zifeng Wu, Chunhua Shen, and Anton Van Den~Hengel.
\newblock Wider or deeper: Revisiting the resnet model for visual recognition.
\newblock {\em Pattern recognition}, 90:119--133, 2019.

\bibitem{russakovsky2015imagenet}
Olga Russakovsky, Jia Deng, Hao Su, Jonathan Krause, Sanjeev Satheesh, Sean Ma, Zhiheng Huang, Andrej Karpathy, Aditya Khosla, Michael Bernstein, et~al.
\newblock Imagenet large scale visual recognition challenge.
\newblock {\em International journal of computer vision}, 115:211--252, 2015.

\bibitem{simonyan2014very}
Karen Simonyan and Andrew Zisserman.
\newblock Very deep convolutional networks for large-scale image recognition.
\newblock {\em arXiv preprint arXiv:1409.1556}, 2014.

\bibitem{achanta2012slic}
Radhakrishna Achanta, Appu Shaji, Kevin Smith, Aurelien Lucchi, Pascal Fua, and Sabine S{\"u}sstrunk.
\newblock Slic superpixels compared to state-of-the-art superpixel methods.
\newblock {\em IEEE transactions on pattern analysis and machine intelligence}, 34(11):2274--2282, 2012.

\bibitem{zhao2024sfc}
Xinqiao Zhao, Feilong Tang, Xiaoyang Wang, and Jimin Xiao.
\newblock \uppercase{Sfc}: Shared feature calibration in weakly supervised semantic segmentation.
\newblock In {\em Proceedings of the AAAI Conference on Artificial Intelligence}, volume~38, pages 7525--7533, 2024.

\bibitem{kweon2023weakly}
Hyeokjun Kweon, Sung-Hoon Yoon, and Kuk-Jin Yoon.
\newblock Weakly supervised semantic segmentation via adversarial learning of classifier and reconstructor.
\newblock In {\em Proceedings of the IEEE/CVF conference on computer vision and pattern recognition}, pages 11329--11339, 2023.

\bibitem{xu2022multi}
Lian Xu, Wanli Ouyang, Mohammed Bennamoun, Farid Boussaid, and Dan Xu.
\newblock Multi-class token transformer for weakly supervised semantic segmentation.
\newblock In {\em Proceedings of the IEEE/CVF conference on computer vision and pattern recognition}, pages 4310--4319, 2022.

\bibitem{lin2014microsoft}
Tsung-Yi Lin, Michael Maire, Serge Belongie, James Hays, Pietro Perona, Deva Ramanan, Piotr Doll{\'a}r, and C~Lawrence Zitnick.
\newblock Microsoft \uppercase{C}o\uppercase{C}o: Common objects in context.
\newblock In {\em Computer Vision--ECCV 2014: 13th European Conference, Zurich, Switzerland, September 6-12, 2014, Proceedings, Part V 13}, pages 740--755. Springer, 2014.

\bibitem{lee2021railroad}
Seungho Lee, Minhyun Lee, Jongwuk Lee, and Hyunjung Shim.
\newblock Railroad is not a train: Saliency as pseudo-pixel supervision for weakly supervised semantic segmentation.
\newblock In {\em Proceedings of the IEEE/CVF conference on computer vision and pattern recognition}, pages 5495--5505, 2021.

\bibitem{sun2021ecs}
Kunyang Sun, Haoqing Shi, Zhengming Zhang, and Yongming Huang.
\newblock \uppercase{Ecs-n}et: Improving weakly supervised semantic segmentation by using connections between class activation maps.
\newblock In {\em Proceedings of the IEEE/CVF international conference on computer vision}, pages 7283--7292, 2021.

\bibitem{chen2024spatial}
Tao Chen, Yazhou Yao, Xingguo Huang, Zechao Li, Liqiang Nie, and Jinhui Tang.
\newblock Spatial structure constraints for weakly supervised semantic segmentation.
\newblock {\em IEEE Transactions on Image Processing}, 2024.

\bibitem{meng2019weakly}
Fanman Meng, Kunming Luo, Hongliang Li, Qingbo Wu, and Xiaolong Xu.
\newblock Weakly supervised semantic segmentation by a class-level multiple group cosegmentation and foreground fusion strategy.
\newblock {\em IEEE Transactions on Circuits and Systems for Video Technology}, 30(12):4823--4836, 2019.

\bibitem{badrinarayanan2017segnet}
Vijay Badrinarayanan, Alex Kendall, and Roberto Cipolla.
\newblock Seg\uppercase{n}et: A deep convolutional encoder-decoder architecture for image segmentation.
\newblock {\em IEEE transactions on pattern analysis and machine intelligence}, 39(12):2481--2495, 2017.

\bibitem{pytorch}
Adam Paszke, Sam Gross, Francisco Massa, Adam Lerer, James Bradbury, Gregory Chanan, Trevor Killeen, Zeming Lin, Natalia Gimelshein, Luca Antiga, et~al.
\newblock Pytorch: An imperative style, high-performance deep learning library.
\newblock In {\em Advances in Neural Information Processing Systems}, volume~32, pages 8024--8035, 2019.

\bibitem{zou2023object}
Zhengxia Zou, Keyan Chen, Zhenwei Shi, Yuhong Guo, and Jieping Ye.
\newblock Object detection in 20 years: A survey.
\newblock {\em Proceedings of the IEEE}, 111(3):257--276, 2023.

\bibitem{zeiler2014visualizing}
Matthew~D Zeiler and Rob Fergus.
\newblock Visualizing and understanding convolutional networks.
\newblock In {\em European conference on computer vision}, pages 818--833. Springer, 2014.

\end{thebibliography}

\end{document}